\documentclass{article}

\usepackage{arxiv}

\usepackage[utf8]{inputenc} 
\usepackage[T1]{fontenc}    
\usepackage{hyperref}       
\usepackage{url}            
\usepackage{booktabs}       
\usepackage{amsfonts}       
\usepackage{nicefrac}       
\usepackage{microtype}      
\usepackage{lipsum}		
\usepackage{graphicx}
\usepackage[authoryear,round]{natbib}
\usepackage{doi}
\usepackage{bm}
\usepackage{mathrsfs}%
\usepackage{amsmath}
\usepackage{amssymb}
\usepackage{tabularx}

\title{Breaking survivorship bias in driving world models}

\author{
  Zihao Sheng$^{1,2}$ \quad
  Zilin Huang$^{1,2}$ \quad
  Yuhao Luo$^{1,2}$ \quad
  Zhengyang Wan$^{1,2}$ \quad
  Boyue Wang$^{1,2}$ \\
  \bfseries
  Soyoung Ahn$^{1,2}$ \quad
  David A. Noyce$^{1,2}$ \quad
  Sikai Chen$^{1,2*}$ \\[1em]
  \normalfont\normalsize
  $^{1}$Department of Civil and Environmental Engineering,\\
  University of Wisconsin--Madison, Madison, WI, USA\\[0.3em]
  $^{2}$Wisconsin Transportation Center,\\
  University of Wisconsin--Madison, Madison, WI, USA \\[0.3em]
  *Corresponding author: \texttt{sikai.chen@wisc.edu}
}

\renewcommand{\shorttitle}{Neutral World Model}

\begin{document}
\maketitle

\begin{abstract}
    World models allow autonomous vehicles to imagine the consequences of candidate actions before execution, but their reliability depends on the futures represented in their training data.
    We identify a \textit{\textbf{survivorship bias}} in driving world models: because the futures recorded in prevailing driving datasets almost always end safely, existing models often imagine benign outcomes even when a collision is imminent.
    We introduce a neutral world model (NWM) that learns both routine and safety-critical futures from a mixture of routine driving and real accident videos with a bounding-box-guided objective.
    The NWM more faithfully anticipates imminent collisions, while maintaining a low false collision rate and competitive generation quality in routine driving.
    We further use the model as an imagination engine for decision-making.
    Given multiple ego trajectories, the NWM predicts the future associated with each candidate, and a reward function evaluates these imagined outcomes before selecting an action.
    On challenging NAVSIM scenarios, this procedure selects the better trajectory more reliably than the baseline world model.
    Zero-shot experiments with real-world pedestrian dummy crossing data collected using an autonomous research vehicle further demonstrate transfer to unseen sensor observations.
    These results show that correcting survivorship bias enables world models to more faithfully reason about futures and support safety-critical decision-making in autonomous driving.
\end{abstract}


\textbf{\textit{Website}}: \url{https://zihaosheng.github.io/NWM/}

\section{Introduction}\label{sec:intro}

Autonomous driving is undergoing a paradigm shift. Instead of decomposing autonomy into separate modules for perception, prediction, and planning, recent research increasingly models driving as an end-to-end process directly from raw sensor observations to driving actions~\citep{hwang2024emma,jiang2025survey,zhang2025pure}. 
At the center of this shift is the emergence of driving world models: generative AI models that predict how future driving scenes evolve given the current state and potential driving actions~\citep{ha2018recurrent,wang2024driving,chen2024end,ding2025understanding,waymo_world_model_2026}.
By enabling an autonomous agent to ``imagine" possible futures before acting, world models are widely viewed as key infrastructure for next-generation autonomous driving~\citep{ren2025cosmos,zheng2026x}.
Yet this promise rests on a critical assumption: that the worlds learned from real driving data are sufficiently representative of the futures an autonomous vehicle must reason about.

\begin{figure}[!t]
    \centering
    \includegraphics[width=0.99\linewidth]{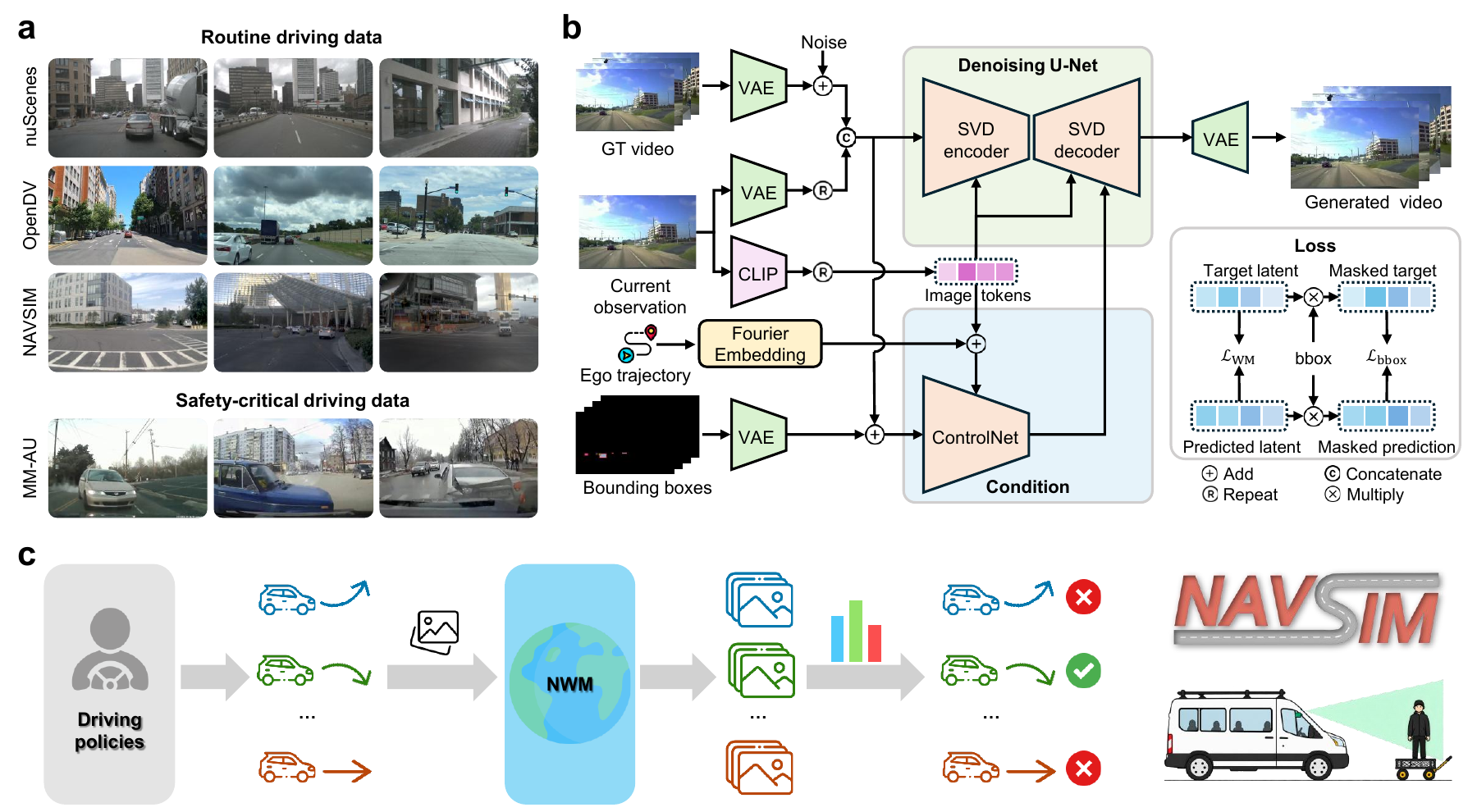}
    \caption{\textbf{Overview of the neutral world model.}
    \textbf{a}, Representative frames from routine driving datasets (nuScenes, OpenDV and NAVSIM) and from the accident dataset MM-AU. The neutral world model (NWM) is trained on both types of data (nuScenes and MM-AU).
    \textbf{b}, Architecture of the NWM, a video diffusion model conditioned on the initial frame, the ego trajectory and optional bounding boxes of surrounding agents. The bounding-box-guided loss $\mathcal{L}_{\mathrm{bbox}}$ complements the diffusion loss $\mathcal{L}_{\mathrm{WM}}$ by emphasizing regions occupied by these agents.
    \textbf{c}, World-model-guided action selection. The NWM imagines the future for each of $K$ candidate ego trajectories, and the candidate with the highest reward is selected. The procedure is evaluated on NAVSIM and on real-vehicle tests with a pedestrian dummy.}
    \label{fig:overview}
\end{figure}

However, the data underlying these models suffer from a fundamental limitation. 
Prevailing driving datasets, such as nuScenes~\citep{caesar2020nuscenes}, OpenDV~\citep{yang2024generalized}, and NAVSIM~\citep{dauner2024navsim}, consist almost exclusively of nominal, collision-free driving (see Fig.~\ref{fig:overview}a). 
This skewed composition is inherent to real-world data collection. Safety-critical events are rare on public roads, many cannot be deliberately collected at scale without unacceptable risk~\citep{yang2023unisim,liu2024curse}, and the drivers and safety operators who record these data resolve most conflicts before they escalate. 
As a result, these datasets do not merely contain few collisions: when a risky interaction does appear in a log, its recorded future almost always ends safely. A world model trained on such data therefore learns what follows a dangerous situation only from the interactions that survived it. 
This constitutes a \textbf{\textit{survivorship bias}} in driving world models: the data contain almost exclusively interactions that ended safely, so the outcomes left out are precisely the dangerous ones that matter most for safety.

This bias leads to systematic failures when world models face safety-critical situations.
We test this by conditioning state-of-the-art driving world models (i.e., Vista~\citep{gao2024vista} and GEM~\citep{hassan2025gem}) on dashcam footage recorded shortly before real traffic accidents~\citep{fang2024abductive} (Section~\ref{sec:survivorship_bias}).
In reality, every one of these clips ends in a collision.
Yet these models often imagine a future in which the collision never happens.
Rather than following the scene dynamics, they reproduce the safe endings that dominate their training data.
These results suggest that existing world models struggle to predict how collisions unfold.
Instead, they have acquired an optimistic prior that favors benign outcomes.

This optimism is especially harmful when world models are used to choose actions.
If a world model cannot imagine a collision, a dangerous trajectory may appear as safe as a cautious one, and imagining before acting offers little protection.
The root of this problem lies in the scarcity of safety-critical data~\citep{xu2022safebench,ding2023survey}.
A natural remedy is to supply more such data.
Because collisions are rare in the logs of data-collection vehicles, many methods synthesize adversarial scenarios instead~\citep{wang2021advsim,feng2021intelligent,rempe2022generating,hanselmann2022king,feng2023dense}.
However, most of them output abstract representations, such as trajectories or agent states, rather than camera images~\citep{rempe2022generating,feng2022trafficgen}.
Such representations lack the visual supervision a video world model needs to learn how dangerous events unfold.
Recent video diffusion models can synthesize realistic crash videos~\citep{li2025avd2,gosselin2025ctrl,li2025causal}. However, they are not conditioned on the ego trajectory and thus cannot imagine the consequences of candidate ego actions.
Consequently, the survivorship bias in driving world models remains unresolved.

Here we propose a framework to break this bias.
We first develop a neutral world model (NWM), which faithfully imagines both nominal and safety-critical driving outcomes without bias towards either.
It is trained on real accident videos alongside routine driving data, which provide visual supervision of dangerous events that trajectory-level scenario representations lack.
As shown in Fig.~\ref{fig:overview}b, a bounding-box-guided loss further emphasizes the regions occupied by surrounding road users, improving the modeling of safety-relevant agent dynamics while preserving generation quality in routine driving.
We then use the NWM as an imagination engine for decision-making (Fig.~\ref{fig:overview}c).
Given multiple candidate ego trajectories proposed by driving policies, the NWM imagines the future that each trajectory would lead to.
A reward function then scores these imagined futures and selects the best trajectory for execution.
We evaluate this approach on challenging scenarios from NAVSIM, in which two state-of-the-art driving policies disagree and only one of them succeeds.
This setting tests whether the imagined futures can reliably identify the better candidate trajectory.
Finally, we conduct zero-shot real-world validation using a lab-developed autonomous research van.
We collect real sensor data in controlled safety-critical scenarios involving staged pedestrian crossings to assess the real-world generalizability of world-model-guided action selection.

We report three principal results.
First, the NWM faithfully imagines collisions in safety-critical scenarios where existing world models tend to generate overly optimistic futures.
At the same time, it maintains a relatively low false collision rate and competitive generation quality in routine driving.
Second, on challenging NAVSIM scenarios where state-of-the-art driving policies disagree, NWM-guided action selection favors better trajectories more reliably than selection guided by Vista.
Third, zero-shot experiments on real-world pedestrian crossing scenarios show that this capability generalizes beyond benchmark data to real sensor observations.
Together, these results demonstrate that breaking survivorship bias not only makes future prediction more faithful but also directly improves action selection in safety-critical driving.

\section{Methods}

\subsection{Neutral world model}
\label{sec:nwm}

\subsubsection{Model architecture and conditioning}

We formulate the neutral world model (NWM) as a controllable latent video diffusion model for future scene generation (Fig.~\ref{fig:overview}b).
Given an initial observation, a candidate ego trajectory, and optional agent-level spatial controls, the model generates a sequence of future camera frames.
The model builds on a pretrained Stable Video Diffusion (SVD) backbone~\citep{blattmann2023stable} with a ControlNet adapter~\citep{zhang2023adding}.
Unlike conventional driving world models trained predominantly on nominal driving, the NWM is trained on both routine driving and real accident videos, as detailed in Section~\ref{sec:data}.
Training follows a two-stage procedure.
In the first stage, we fine-tune the SVD backbone on driving videos to adapt it to the driving domain.
In the second stage, we freeze the backbone and train the ControlNet branch to incorporate the conditioning signals.

Let $\mathbf{x}_{1:T}$ denote a video clip and $\mathbf{z}_{1:T}$ its latent representation obtained with a pretrained variational autoencoder (VAE)~\citep{kingma2013auto}.
Noise is added to the latent according to the diffusion schedule~\citep{ho2020denoising}, and the network is trained to recover the clean latent:
\begin{equation}
\mathcal{L}_{\mathrm{wm}} = \mathbb{E}\Big[ \lambda(s) \big\lVert \hat{\mathbf{z}}_{1:T} - \mathbf{z}_{1:T} \big\rVert_2^2 \Big],
\label{eq:wm}
\end{equation}
where $\hat{\mathbf{z}}_{1:T}$ is the denoised latent, $s$ is the diffusion timestep and $\lambda(s)$ is the loss weighting.

The model receives three types of conditioning input:
\begin{itemize}
\item \textbf{Visual context.} The current camera observation provides the visual context, including scene layout, lighting, road geometry and the appearance of surrounding agents. It is encoded by the VAE and a CLIP image encoder~\citep{radford2021learning}.
\item \textbf{Ego trajectory.} The ego trajectory is a sequence of future poses $\tau^{\mathrm{ego}}_{1:T} = (x_t, y_t, \psi_t)_{t=1}^{T}$, where $(x_t, y_t)$ is the position and $\psi_t$ the yaw angle. The positions are encoded with sinusoidal Fourier features~\citep{tancik2020fourier,gao2024vista} and added to the image embedding.
The resulting feature is then fed into the cross-attention layers of ControlNet. This allows the model to generate futures consistent with the intended ego motion.
\item \textbf{Agent bounding boxes.} Surrounding road users are represented as 2D bounding boxes encoding their positions, sizes and identities. At each timestep, the boxes are rasterized into an RGB control frame aligned with the camera view, encoded by the VAE and processed by the ControlNet branch. During training, we sample a random time step and replace all controls from that step onward with a learnable null embedding. The model thus learns to generate futures with full, partial or no agent controls. At inference, unavailable controls are represented in the same way.
\end{itemize}

\subsubsection{Bounding-box-guided loss}

In safety-critical scenes, the most important changes often occupy a small image region but evolve rapidly.
The objective in Equation~\eqref{eq:wm} weights all latent locations equally and is therefore dominated by background and global appearance.
Under this objective, the model generates high-quality nominal videos but models the behavior of safety-relevant agents poorly.
In the second training stage, we therefore augment Equation~\eqref{eq:wm} with a bounding-box-guided loss that increases the training weight on latent regions associated with surrounding agents (see $\mathcal{L}_{\mathrm{bbox}}$ in Fig.~\ref{fig:overview}b).

For each frame, we construct a binary mask $M_t$ from the rasterized bounding boxes.
We encode the mask sequence and an all-zero sequence with the same VAE to obtain latents $\mathbf{z}^{M}_{1:T}$ and $\mathbf{z}^{\varnothing}_{1:T}$, and compute
\begin{equation}
A_{t,c,i,j} = \big( \mathbf{z}^{M}_{t,c,i,j} - \mathbf{z}^{\varnothing}_{t,c,i,j} \big)^2 ,
\label{eq:saliency}
\end{equation}
which maps the box support into the latent space of the diffusion target while cancelling the VAE response to an empty input.
We then apply $\ell_2$ normalization to $A$ over the temporal and spatial dimensions, independently for each latent channel, to obtain $\bar{A}$.
The element-wise loss weight is
\begin{equation}
W_{t,c,i,j} = 1 + q_t\,\bar{A}_{t,c,i,j},
\label{eq:weight}
\end{equation}
where $q_t \in \{0,1\}$ indicates whether the bounding-box condition is available at frame $t$.
The final objective is
\begin{equation}
\mathcal{L}_{\mathrm{neutral}} = \mathbb{E}\Big[ \lambda(s) \sum_{t,c,i,j} W_{t,c,i,j} \big( \hat{\mathbf{z}}_{t,c,i,j} - \mathbf{z}_{t,c,i,j} \big)^2 \Big].
\label{eq:neutral}
\end{equation}

This formulation has two useful properties.
First, the weights are derived from the data rather than from hand-designed heuristics, so they adapt to varying agent sizes and occlusion patterns.
Second, the additive form of Equation~\eqref{eq:weight} preserves the original objective.
Every latent element keeps a base weight of one, and box-sensitive elements receive additional emphasis.
Together with the accident videos, this loss helps the model to learn the motions of surrounding agents, including collisions and near misses, while retaining its capacity to generate routine driving.

\subsubsection{Controllable future generation}

At inference, the NWM generates a future video as
\begin{equation}
\hat{\mathbf{x}}_{1:T} = G_\theta\big(\mathbf{o}, \tau^{\mathrm{ego}}_{1:T}, \mathcal{B}_{1:T}\big),
\label{eq:generation}
\end{equation}
where $G_\theta$ denotes the trained model, $\mathbf{o}$ is the current observation and $\mathcal{B}_{1:T}$ is the set of bounding-box sequences of the controlled agents, which may be empty.
This interface supports two modes of generation.

With agent controls, we specify future bounding-box trajectories for selected agents.
These trajectories may describe nominal continuations, near misses or collision courses.
When trajectories are edited in three-dimensional space, we project them into the image plane using calibrated camera intrinsics and extrinsics before rasterization.
This projection keeps the boxes geometrically consistent with the scene, preserving realistic scaling, foreshortening and occlusion as agents approach, collide or pass at close range.
In collision cases, the generated videos can exhibit physically plausible secondary effects, such as vehicle deformation and camera shake.
We use this mode to evaluate controllability (Section~2.3).

Without agent controls ($\mathcal{B}_{1:T} = \emptyset$), no future agent motion is given.
The model must then infer from the observation how surrounding agents will move and whether they will come into conflict with the ego vehicle.
We use this mode to quantify survivorship bias (Section~\ref{sec:survivorship_bias}) and for action selection (Section~\ref{sec:selection}).
In both cases, the future motions of other agents are not given and must be imagined by the model.

\subsection{World-model-guided action selection}
\label{sec:selection}

\subsubsection{Candidate trajectories and future imagination}

At each decision step, we consider a set of $K$ candidate ego trajectories $\{\tau^{(k)}_{1:T}\}_{k=1}^{K}$ (Fig.~\ref{fig:overview}c).
The selection procedure is agnostic to how the candidates are produced.
For each candidate, the NWM imagines the resulting future without agent controls:
\begin{equation}
\hat{\mathbf{x}}^{(k)}_{1:T} = G_\theta\big(\mathbf{o}, \tau^{(k)}_{1:T}, \emptyset\big), \quad k = 1, \dots, K.
\label{eq:imagination}
\end{equation}
All candidates share the same current observation and differ only in the ego trajectory.
Differences among their imagined futures therefore reflect the consequences of the candidate actions.

\subsubsection{Reward-based trajectory selection}

Each imagined future is scored by a reward function that accounts for safety, progress and comfort:
\begin{equation}
R\big(\hat{\mathbf{x}}^{(k)}, \tau^{(k)}\big) = S \cdot \frac{w_{\mathrm{T}}\,\mathrm{TTC} + w_{\mathrm{P}}\,P + w_{\mathrm{C}}\,C}{w_{\mathrm{T}} + w_{\mathrm{P}} + w_{\mathrm{C}}},
\label{eq:reward}
\end{equation}
where $S$ is a safety gate, $\mathrm{TTC}$ is a temporal safety margin, $P$ is a progress term and $C$ is a comfort term, each taking values in $[0,1]$, and $w_{\mathrm{T}}$, $w_{\mathrm{P}}$ and $w_{\mathrm{C}}$ are their weights.
This form follows the structure of the PDM score used in NAVSIM~\citep{dauner2024navsim}.

The two safety terms play complementary roles.
The safety gate $S$ penalizes imagined collisions, and penalizes them more strongly when they occur earlier.
The multiplicative safety gate reduces the contribution of progress and comfort when a collision is predicted.
The temporal margin $\mathrm{TTC}$ measures how long the ego vehicle keeps a safe clearance from other road users.
It distinguishes a near miss from a comfortable pass, even when neither ends in a collision.

The progress term $P$ measures how appropriately a candidate advances along its path.
What counts as appropriate progress depends on the surrounding traffic.
We therefore condition $P$ on a traffic context that the NWM infers from the current state.
The same context is used for all candidates from that state, so they are compared on equal terms.
Like the safety terms, $P$ thus depends on what the world model imagines rather than on ground-truth information.
The comfort term $C$ penalizes large acceleration and jerk, discouraging harsh maneuvers.
The candidate with the highest reward is selected:
\begin{equation}
k^{*} = \arg\max_{k} R\big(\hat{\mathbf{x}}^{(k)}, \tau^{(k)}\big).
\label{eq:select}
\end{equation}
The detailed reward computation is described in Supplementary Sections 5.1–5.4, Supplementary Fig. 11 and Supplementary Table 5.

\section{Results}\label{sec2}

\subsection{Datasets and experimental setup}

\subsubsection{Training data}\label{sec:data}
The NWM is trained on nuScenes~\citep{caesar2020nuscenes} and MM-AU~\citep{fang2024abductive}.
nuScenes provides routine urban driving with calibrated cameras, ego poses and 3D road-user annotations.
MM-AU provides egocentric accident videos, from which we extract both crash and nominal clips.
All samples are converted into 25-frame front-view clips at $512 \times 320$ resolution.
The two datasets play complementary roles: nuScenes anchors the model in routine driving, whereas MM-AU provides direct visual supervision for accident dynamics that are largely absent from conventional driving datasets.
For MM-AU, ego trajectories are estimated using a depth-assisted pseudo-labeling pipeline (Supplementary Sections 2.4–2.5 and Supplementary Fig. 2).

For nuScenes, we use the official training and validation splits of 700 and 150 scenes.
Overlapping sampling with a stride of five frames yields 7,700 training clips, and non-overlapping sampling yields 450 validation clips.
For MM-AU, we start from the predefined training and validation splits of 8,815 and 1,022 videos.
We remove videos that fail quality filtering, have fewer than 25 frames, or lack the required annotations.
From each remaining video, we extract a crash clip. When a sufficiently long pre-event interval is available, we additionally extract a nominal clip from the earliest window.
This yields 7,939 training clips (6,194 crash and 1,745 nominal) and 952 validation clips (733 crash and 219 nominal).

To counter the dominance of straight driving, we repeat left- and right-turn clips five times.
After this trajectory-based balancing, MM-AU and nuScenes contribute 12,011 and 11,500 effective training samples, respectively, corresponding to a dataset ratio of 51.1\% to 48.9\%. The resulting right-turn, left-turn, stopping, and straight-driving proportions are 23.3\%, 18.6\%, 21.4\%, and 36.7\%, respectively. 
Detailed dataset statistics and the effect of maneuver-based balancing are provided in Supplementary Table 3 and Supplementary Fig. 1.

\subsubsection{Survivorship-bias evaluation}
To quantify survivorship bias, we select 100 ego-involved accident scenes from the MM-AU validation split.
For each scene, we construct four prediction windows beginning approximately $2.0$, $1.5$, $1.0$, and $0.5$~s before the annotated collision. 
At inference, the model is conditioned on the initial RGB frame and ego trajectory, and no future agent motion is specified.
The model must therefore infer from the visible interaction whether a collision will occur, rather than follow prescribed future boxes.
Results are compared with the reference accident clips.

\subsubsection{Routine driving evaluation}
To test whether the NWM imagines collisions where none occur, we use 50 collision-free driving clips from MM-AU and 100 clips from nuScenes~\citep{caesar2020nuscenes}.
The MM-AU clips share the visual domain of the accident videos.
They therefore test whether the model has merely learned to reproduce accidents from the appearance of MM-AU footage.
The same clips are used to assess generation quality.
Models are conditioned in the same way as in the survivorship-bias evaluation.

\subsubsection{Controllability evaluation}
To evaluate agent-level control under motions that were never recorded, we additionally synthesize adversarial and collision-avoidance trajectories with STRIVE~\citep{rempe2022generating} for 158 scenes from nuScenes. 
The scenes cover two categories of conflict, including vehicle-involved and pedestrian-involved.
Each trajectory is combined with surrounding agents' dimensions to form oriented 3D bounding boxes, which are projected and rasterized as complete bounding-box control sequences. 
Unlike the survivorship-bias evaluation, the dangerous motion here is prescribed rather than imagined, so this experiment tests control fidelity rather than collision anticipation.

\subsubsection{NAVSIM action selection}
We evaluate action selection on a challenging subset of NAVSIM. 
For each scene, two end-to-end policies (ExploreVLA~\citep{sheng2026explorevla} and AutoVLA~\citep{zhou2025autovla}) produce candidate trajectories from the same sensor observation. 
We then select a subset of 200 challenging scenes in which only one candidate succeeds under the NAVSIM evaluator.
Each candidate is converted to the NWM trajectory representation, imagined independently, scored, and selected as in Equation~\eqref{eq:select}. 

\subsubsection{Real-world zero-shot data}

Real-world data were collected using a Ford E-Transit electric research van operated by the Wisconsin Transportation Center at the University of Wisconsin--Madison. The experiments were conducted on a controlled route in Fitchburg, Wisconsin, USA, on February 23 and March 17, 2026. 
A human safety operator drove the vehicle at low speed while synchronized front-camera images and ego-state measurements were recorded. The NWM was not fine-tuned on these data. All field results therefore constitute zero-shot transfer of the learned video model and action-selection procedure.

A pedestrian dummy mounted on a wheeled cart was moved across the vehicle path to create repeatable conflict events without placing a participant in the ego lane. The recordings contain approach, deceleration, and stopped states.
We selected decision frames at two stages of each trial: before the crossing, when the vehicle was driving normally, and during the vehicle's response, when the safety operator was braking for the dummy.
At each decision frame, the candidate set consists of two ego trajectories: the trajectory actually driven by the safety operator, recorded from the ego-state measurements, and a counterfactual alternative.
Before the crossing, the alternative is to brake. During the response, the alternative is to resume driving.
The driven trajectory therefore serves as the correct choice (Section~\ref{sec:metrics}), and selecting the alternative corresponds to unnecessary braking or a failure to stop for the pedestrian, respectively.
Experimental variation includes dry and snowy conditions, changes in the dummy's clothing, and visible and occluded crossings in which the dummy emerged from behind a parked vehicle.

\subsubsection{Baselines}
We compare the NWM with Vista~\citep{gao2024vista} and GEM~\citep{hassan2025gem} for routine and accident future generation.
Each baseline receives the same observation and ego trajectory as the NWM.
In action selection, Vista serves as the world-model baseline and uses the same reward (Section~\ref{sec:selection}).
All videos are evaluated at a common $512\times320$ resolution and over the same 25 frames.

\subsection{Evaluation metrics}
\label{sec:metrics}

\subsubsection{Generation quality}\label{sec:generation_quality}
We measure video quality using Fr\'echet Inception Distance (FID)~\citep{heusel2017gans}, Fr\'echet Video Distance (FVD)~\citep{unterthiner2018towards}, and JEDi~\citep{luo2025beyond}. Before feature extraction, videos from all models and their paired references are resized to $512\times320$ using the same interpolation procedure. 
FID evaluates frame-level appearance, whereas FVD and JEDi evaluate spatiotemporal realism. 
Lower FID, FVD, and JEDi scores indicate greater similarity between the generated and real video distributions.

\subsubsection{Outcome fidelity}
We determine the outcome of each generated video by manual inspection, labeling it as either \emph{collision} or \emph{no collision}.
On accident scenes, collision recall is the fraction of inputs whose imagined future contains a collision, reported separately for each prediction starting point.
On nominal clips, the false collision rate is the fraction of inputs whose imagined future contains a collision.

To understand how a world model avoids a collision that should occur, we further examine the accident scenes at the shortest horizon of 0.5\,s.
We assign each video labeled \emph{no collision} at this horizon to one of four avoidance modes:
(i) \emph{agent deviates}, in which the other road user changes its speed or direction to avoid collision;
(ii) \emph{agent vanishes}, in which the other road user fades out of view;
(iii) \emph{ego deviates}, in which the ego motion departs from the conditioned trajectory;
and (iv) \emph{same-speed motion / scene freezes}, in which the ego vehicle and other road users move at similar speeds or the video stops evolving or degrades.

\subsubsection{Controllability}
We measure how closely the generated videos follow the prescribed motion of the controlled agent.
At each frame $t$, the prescribed box $\bm B_t$ is the projection of the agent's edited trajectory into the image.
We keep frames in which the agent is visible and $\bm B_t$ is the largest of the projected control boxes.
In the corresponding generated frame, $\hat{\bm B}_t$ is likewise the largest YOLOv11~\citep{khanam2024yolov11} detection.
For each frame, we compute the intersection over union $\mathrm{IoU}_t$ between $\bm B_t$ and $\hat{\bm B}_t$, and the normalized center offset (nCO),
\begin{equation}
\mathrm{nCO}
=
\frac{1}{T}
\sum_{t=1}^{T}
\frac{
\left\|
\hat{c}_t-c_t
\right\|_2
}{
\sqrt{W_{\mathrm{img}}^2+H_{\mathrm{img}}^2}
},
\label{eq:ncom}
\end{equation}
where $\bm c_t$ and $\hat{\bm c}_t$ are the box centers and $W_{\mathrm{img}} \times H_{\mathrm{img}} = 512 \times 320$ is the image size.
Over all frames in which both boxes are available, we report the mean IoU, the mean nCO and the success rate SR@0.5, defined as the fraction of frames with $\mathrm{IoU}_t \geq 0.5$.
Higher IoU and SR@0.5 and lower nCO indicate closer agreement with the control.

\subsubsection{Action selection}
We evaluate action selection by how often the imagined futures lead to the better of two candidate trajectories.
For each decision sample $i$, let $k_i^{*}$ denote the better candidate according to the ground truth, and let $\hat{k}_i$ denote the candidate selected from the imagined futures (Equation~\eqref{eq:select}).
We define
\begin{equation}
\mathrm{Acc}_{\mathrm{sel}}
=
\frac{1}{N}
\sum_{i=1}^{N}
\mathbb{I}
\left[
\hat{k}_i=k_i^{*}
\right],
\label{eq:selection_acc}
\end{equation}
where $N$ is the number of decision samples and $\mathbb{I}[\cdot]$ is the indicator function.
In NAVSIM, $k_i^{*}$ is the candidate rated higher by the official NAVSIM evaluator.
In the real-world data, $k_i^{*}$ is the trajectory driven by the human safety operator.

\begin{figure}[htbp]
    \centering
    \includegraphics[width=0.99\linewidth]{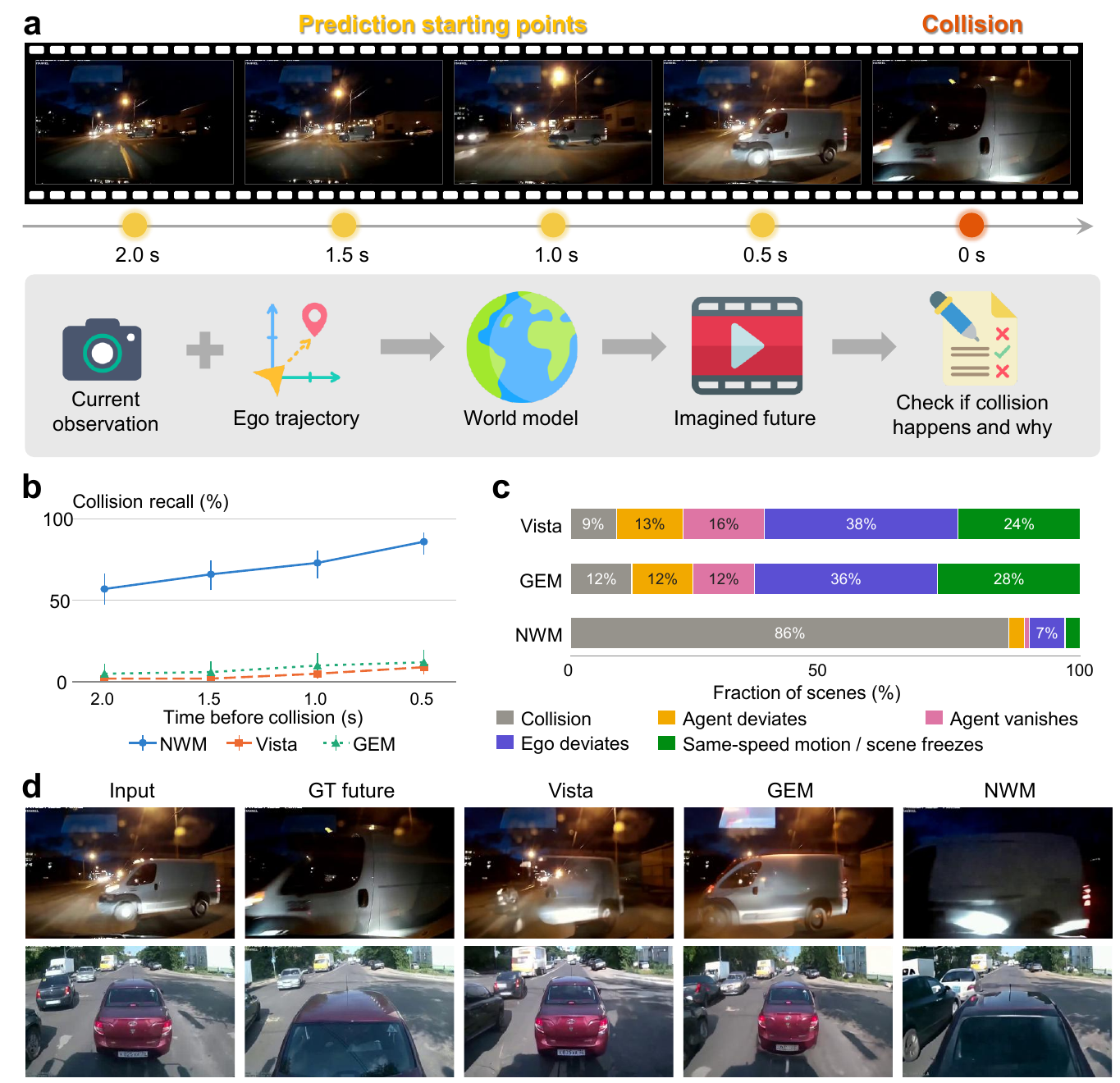}
    \caption{\textbf{Survivorship bias in driving world models.}
    \textbf{a}, Evaluation protocol. For 100 ego-involved accident scenes from MM-AU, prediction starts approximately 2.0, 1.5, 1.0 and 0.5~s before the annotated collision. The world model receives the current observation and the recorded ego trajectory.
    \textbf{b}, Collision recall versus time before collision for NWM, Vista, and GEM ($n = 100$ scenes per point). Error bars show 95\% Wilson score confidence intervals.
    \textbf{c}, Composition of generated outcome at 0.5~s before collision. Each bar partitions the 100 scenes into imagined collisions and four avoidance modes.
    \textbf{d}, Representative examples of a side collision (top) and a rear-end collision (bottom). Vista imagines the van vanishing and the lead car keeping pace with the ego vehicle; GEM imagines the ego vehicle stopping and the lead car accelerating away. The NWM correctly imagines the collision in both.}
    \label{fig:survivorship_bias}
\end{figure}

\subsection{Survivorship bias in driving world models}\label{sec:survivorship_bias}

We first examine whether a driving world model can imagine a collision that is about to happen. 
We select 100 ego-involved accident scenes whose recorded futures all end in a collision and start the prediction at four time points, approximately 2.0, 1.5, 1.0, and 0.5\,s before the annotated collision (Fig.~\ref{fig:survivorship_bias}a).
Each model receives the current observation and the recorded ego trajectory, but no information about how other road users will move. 
Because every reference video ends in a collision, a faithful model should increasingly recover this outcome as the prediction starts closer to the annotated collision. We therefore report collision recall, the fraction of imagined futures that contain a collision.

Existing world models rarely imagine the collision observed in the recorded future (Fig.~\ref{fig:survivorship_bias}b). 
Collision recall increases as the prediction starts closer to collision, but remains low for both Vista and GEM across all four prediction starting points. 
The discrepancy is most pronounced when the collision is already imminent. At only 0.5\,s before collision, Vista and GEM imagine the collision in 9\% and 12\% of the scenes, respectively, whereas our NWM reaches 86\%. 
The persistence of this gap at such a short time before collision indicates that the baseline failures cannot be explained by long-horizon uncertainty alone: even when strong visual evidence of a conflict is present, the models predominantly imagine a non-collision future.

To understand how a collision that occurs in reality is avoided in imagination, we assign each collision-free future at the 0.5~s horizon to one of four avoidance modes (Fig.~\ref{fig:survivorship_bias}c). The dominant failure mode for both Vista and GEM is \textit{ego deviates}, in which the imagined ego vehicle departs from the trajectory on which it was conditioned, for example by slowing or stopping before the conflict. 
Another common mode is \textit{same-speed motion / scene freezes}, in which the relative motion stops: the other vehicle moves in step with the ego vehicle, or the scene stops evolving, so that the gap between them never closes. 
Together, these two modes account for nearly two-thirds of the scenes for both baselines. Other collision-free generations avoid the conflict by altering the behavior of surrounding road users, either changing their motion or making them vanish from the scene.

Representative examples in Fig.~\ref{fig:survivorship_bias}d illustrate this optimistic rewriting of the future. In the first, a van crosses the ego path and the recorded interaction ends in a side collision. Vista avoids the collision by making the van vanish, whereas GEM departs from the conditioned ego trajectory and brings the ego vehicle to a stop. The NWM keeps both vehicles on course and imagines the collision. The second example is a rear-end collision, in which the ego vehicle closes on a lead car until collision. Vista lets the lead car move in step with the ego vehicle, so the gap between them never closes, and GEM has the lead car accelerate away. The NWM again lets the gap close and imagines the collision. Together, these results reveal the survivorship bias in existing driving world models: rather than following the dynamics of a safety-critical interaction, they turn futures that are about to end in a collision into benign ones. For downstream decision-making, the consequence is direct: a world model that imagines a safe ending for a dangerous action cannot distinguish a dangerous candidate trajectory from a cautious one.
Additional visualization generations are shown in Supplementary Fig.~3.

\begin{figure}[!t]
    \centering
    \includegraphics[width=\textwidth]{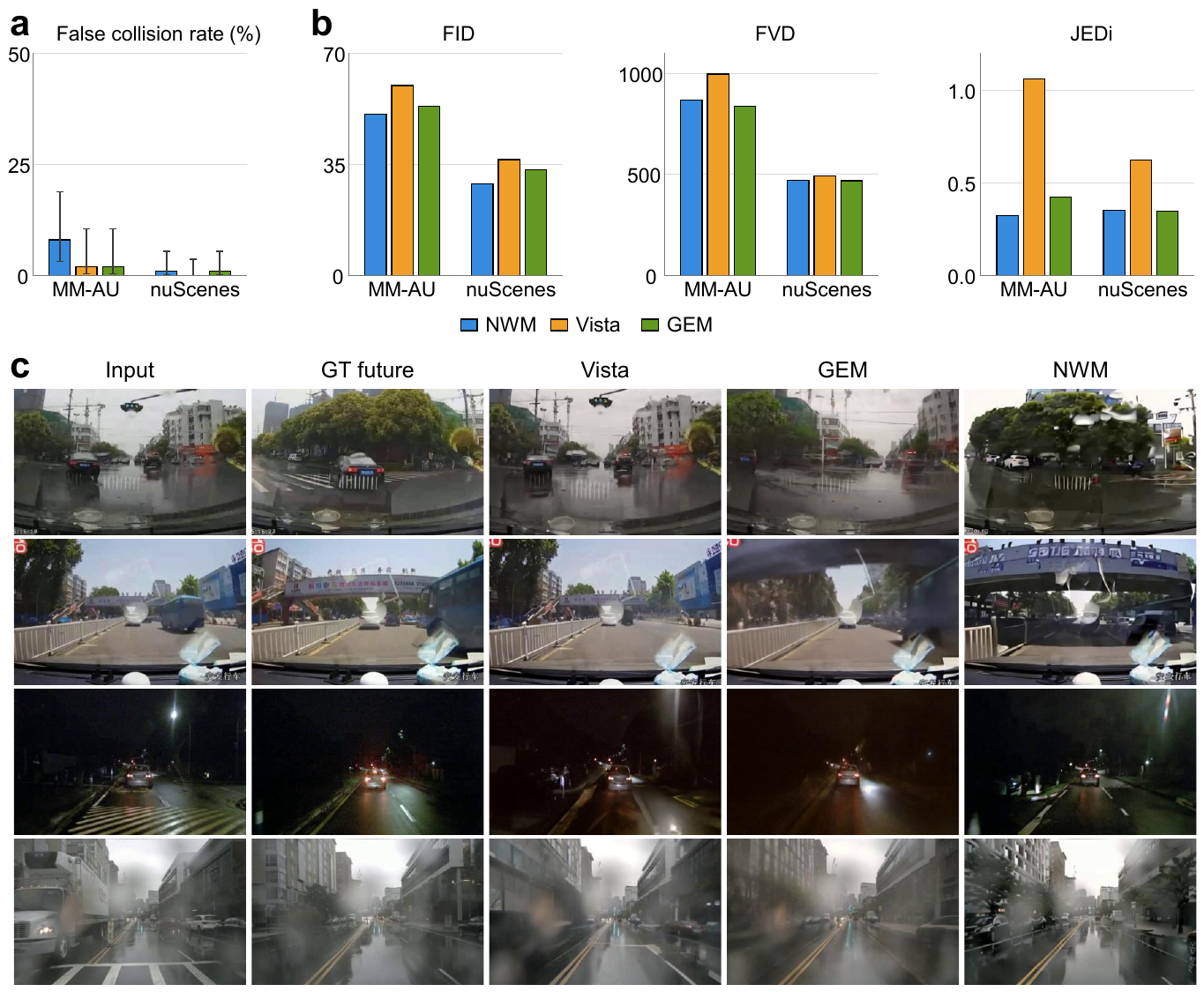}
    \caption{
    \textbf{Correcting survivorship bias preserves routine-driving fidelity.}
    \textbf{a}, False collision rates for NWM, Vista, and GEM on collision-free
    MM-AU ($n=50$) and nuScenes ($n=100$) clips. Error bars show 95\% Wilson
    score confidence intervals.
    \textbf{b}, FID, FVD, and JEDi on routine driving from MM-AU and nuScenes.
    Lower values indicate greater similarity between generated and real videos.
    \textbf{c}, Representative routine-driving generations. Columns show the input
    observation, ground-truth future, and futures generated by Vista, GEM, and NWM.
    The first two rows show an MM-AU turning scene and a daytime car-following
    scene, respectively; the last two rows show a nuScenes nighttime car-following
    scene and a rainy driving scene, respectively.
    }
    \label{fig:routine_generation}
\end{figure}

\subsection{Faithful imagination in routine driving}\label{sec:routine}
Learning to imagine collisions could introduce the opposite failure mode: a world model may become overly pessimistic and hallucinate dangerous outcomes during routine driving. We therefore evaluate all models on collision-free clips from both MM-AU and nuScenes, under the same conditioning as in Section~\ref{sec:survivorship_bias} (Fig.~\ref{fig:routine_generation}). The MM-AU subset is particularly informative because it shares the visual domain of the accident videos used to train the NWM. It therefore tests whether the model has simply associated the appearance of MM-AU footage with accidents. The nuScenes subset provides a complementary evaluation on conventional autonomous-driving data. We report the false collision rate, the fraction of imagined futures that contain a collision.

The false collision rate is low for all three models (Fig.~\ref{fig:routine_generation}a). On the MM-AU clips, the NWM imagines a collision in 8\% of the futures, against 86\% for accident scenes 0.5~s before collision (Fig.~\ref{fig:survivorship_bias}b). For Vista and GEM, the corresponding fractions increase only from 2\% on nominal clips to 9\% and 12\% on accident scenes, respectively, remaining low even when a collision is imminent. On nuScenes, no model exceeds 1\%. Thus, the NWM is more likely to imagine a collision when the interaction is heading for one than in visually similar routine footage. This indicates that its predictions respond to the interaction in the scene rather than to the appearance of MM-AU footage.

This sensitivity to dangerous outcomes does not come at the cost of generation quality in routine driving (Fig.~\ref{fig:routine_generation}b). We assess the similarity between generated and real videos using Fr\'echet Inception Distance (FID)~\citep{heusel2017gans} for frame-level appearance and Fr\'echet Video Distance (FVD)~\citep{unterthiner2018towards} and JEDi~\citep{luo2025beyond} for video-level spatiotemporal content. Lower values indicate greater similarity (Methods Section~\ref{sec:generation_quality}). Across FID, FVD and JEDi, the NWM is comparable to or better than Vista and GEM on both datasets.
The examples in Fig.~\ref{fig:routine_generation}c cover a turn at an intersection, car following by day and at night, and driving in rain. Across these conditions, the NWM produces plausible collision-free futures while preserving the overall scene structure. Together with Section~\ref{sec:survivorship_bias}, these results show that the NWM imagines collisions when they are imminent while rarely hallucinating them during routine driving. This distinction is important for decision-making, where a spurious collision would penalize a safe trajectory just as a missed collision would excuse a dangerous one.
Further examples of routine-driving generations are provided in Supplementary Fig. 4.

\subsection{Agent-level control of safety-critical futures}\label{sec:control}

In Sections~\ref{sec:survivorship_bias} and~\ref{sec:routine}, the NWM is conditioned only on the current observation and the ego trajectory, and it predicts the motion of other road users without further guidance. A complementary capability for counterfactual analysis and safety testing is to generate how the scene would evolve when another road user follows a given motion, including trajectories that were not observed in the recorded data. 
We thus evaluate whether the NWM can faithfully realize a future when that motion is explicitly specified. 
We quantify how closely a generated agent follows its prescribed boxes using the mean intersection over union (mIoU), the fraction of frames with an IoU of at least 0.5 (SR@0.5) and the normalized center offset (nCO) between prescribed and detected boxes.

\begin{figure}[!t]
    \centering
    \includegraphics[width=\textwidth]{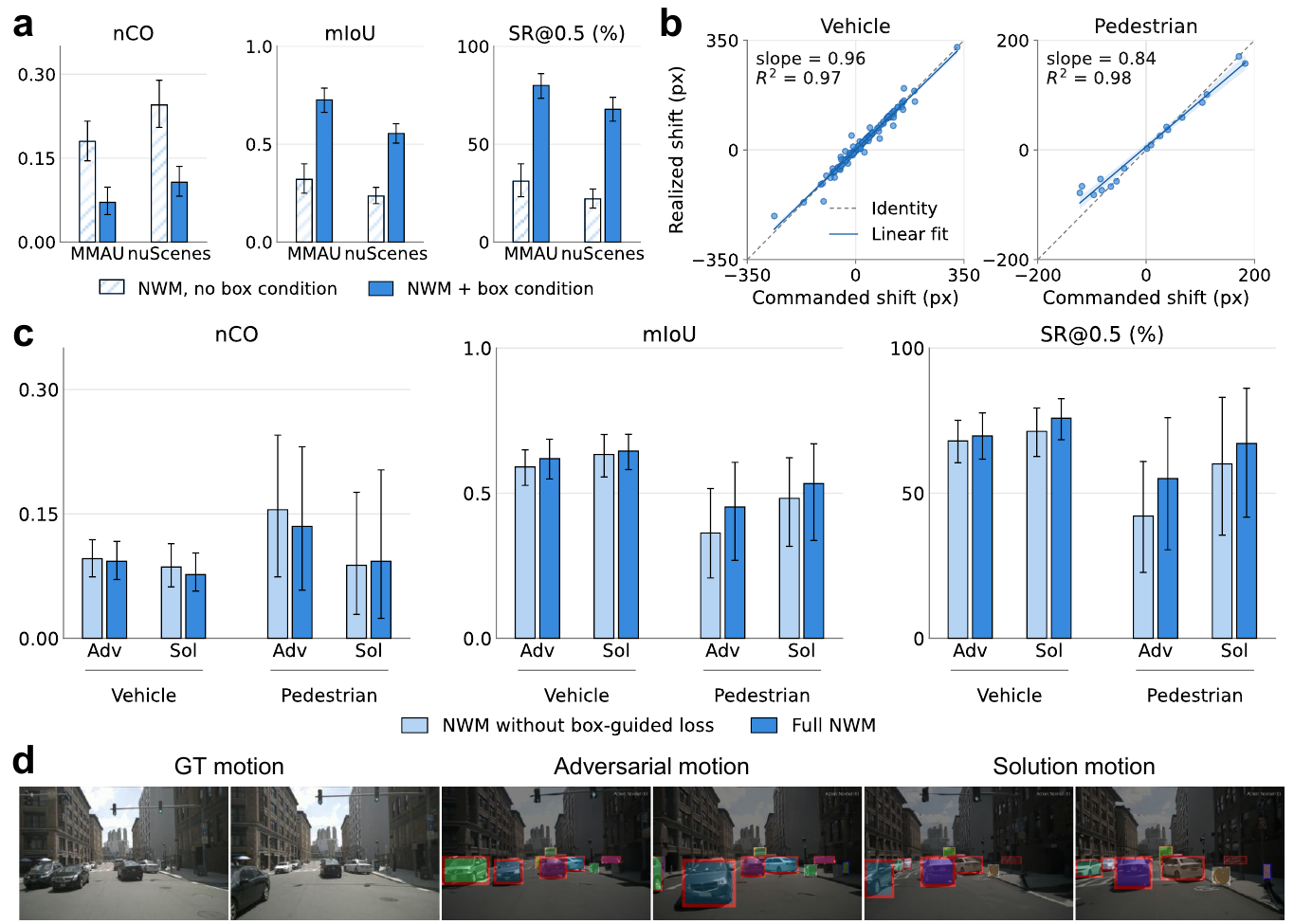}
    \caption{
    \textbf{Agent-level control of safety-critical futures.}
    \textbf{a}, Control fidelity with and without bounding-box conditioning on MM-AU and nuScenes, measured by nCO, mIoU, and SR@0.5.
    \textbf{b}, Prescribed versus realized lateral displacement for vehicles and pedestrians. Dashed lines indicate the identity relation and solid lines show linear fits.
    \textbf{c}, Ablation of the bounding-box-guided loss on STRIVE adversarial (Adv) and solution (Sol) trajectories.
    \textbf{d}, Representative generations under adversarial and solution motions from the same scene. Bounding-box controls are overlaid on the generation. Error bars show 95\% confidence intervals.
    }
    \label{fig:control}
\end{figure}

Bounding-box conditioning substantially improves control fidelity on both MM-AU and nuScenes (Fig.~\ref{fig:control}a). Compared with generation without box controls, conditioning on the prescribed boxes markedly reduces nCO, while increasing mIoU and SR@0.5. The improvement is consistent across the two visual domains.
Explicit spatial controls can thus constrain the generated agents to the prescribed trajectory.

We next consider a more challenging setting in which the prescribed motion departs from the recorded trajectory. For 158 nuScenes scenes, STRIVE~\citep{rempe2022generating} is used to generate an adversarial trajectory that directs another road user into conflict with the ego vehicle, together with a corresponding solution trajectory that avoids the conflict. We consider two categories of ego-involved conflict, in which the adversarial road user is either a vehicle or a pedestrian.
We compare the lateral displacement of the prescribed bounding boxes relative to the recorded GT boxes with the corresponding displacement realized in the generated video (Fig.~\ref{fig:control}b). The two are closely aligned: the fitted slope is 0.96 for vehicles ($R^2=0.97$) and 0.84 for pedestrians ($R^2=0.98$). 
Thus, the realized lateral displacement closely follows the prescribed displacement across a range of counterfactual motions.

To assess the contribution of the bounding-box-guided training objective, we compare the full NWM with a variant without this loss on the STRIVE-edited trajectories (Fig.~\ref{fig:control}c).
Across both adversarial and solution motions, the full NWM achieves higher mIoU and SR@0.5 than the same model trained without the bounding-box-guided loss for both vehicle- and pedestrian-involved scenes. 
The improvement is more pronounced for pedestrian trajectories, which is consistent with the intended emphasis of the loss on smaller safety-relevant regions.

The qualitative example in Fig.~\ref{fig:control}d illustrates how these controls translate into distinct futures from the same driving scene. Under the recorded GT motion, the ego vehicle and surrounding traffic continue normally. Under the adversarial motion, an oncoming vehicle departs from its lane and approaches the ego vehicle, and the generated future follows the prescribed bounding-box motion to produce the resulting conflict. Under the solution motion, the ego vehicle steers to the right to avoid the conflict and the generated view changes accordingly. These results show that the NWM can faithfully represent both nominal and safety-critical counterfactual futures under prescribed motion.
Additional STRIVE-conditioned generations are shown in Supplementary Fig. 7.

\subsection{Imagination-guided decision-making in benchmark and real-world driving}\label{sec:decision}

\begin{figure}[!t]
    \centering
    \includegraphics[width=\textwidth]{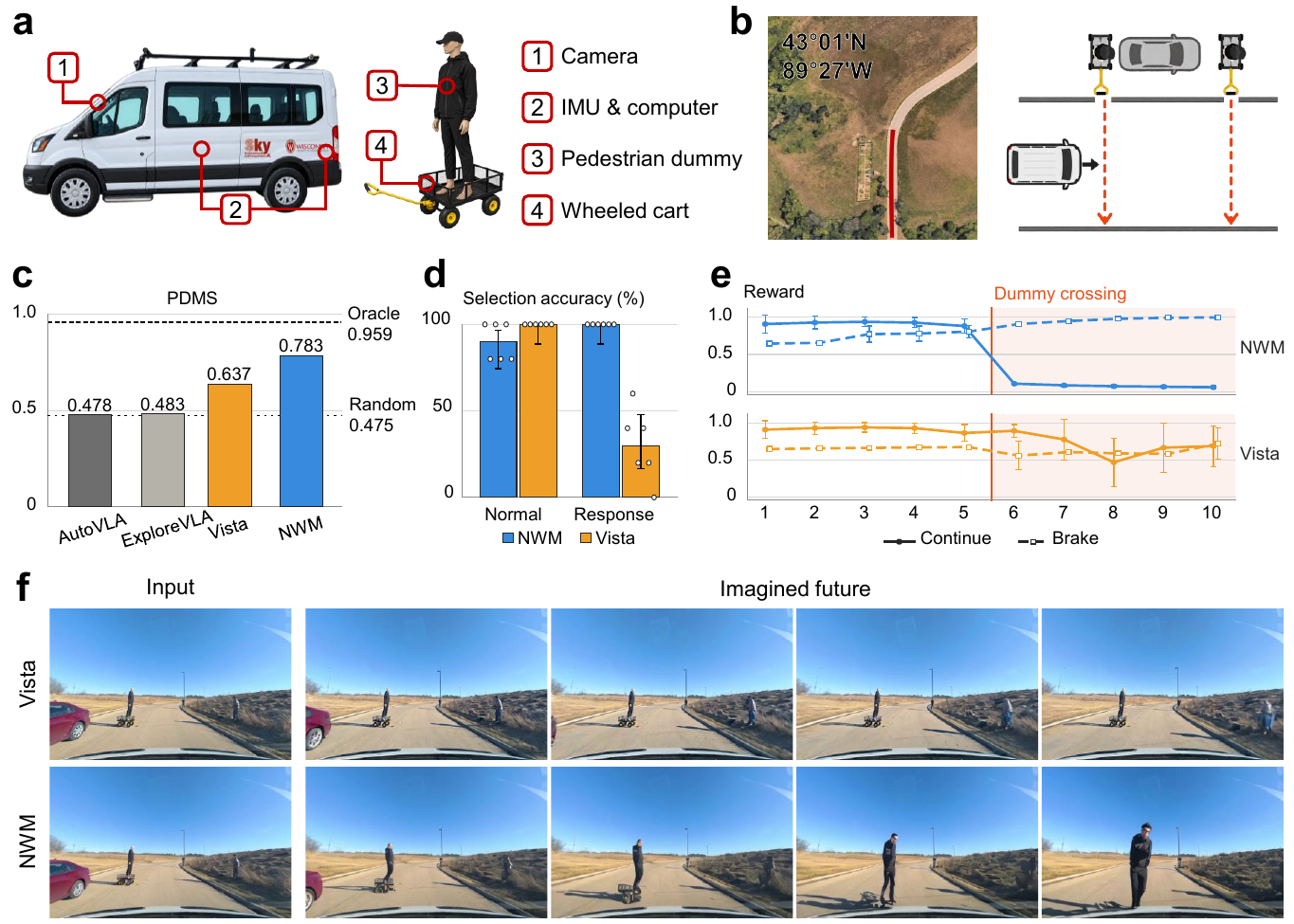}
    \caption{\textbf{Imagination-guided decision-making in benchmark and real-world driving.}
    \textbf{a}, Experimental platform, including the research van operated by the Wisconsin Transportation Center and a pedestrian dummy mounted on a wheeled cart.
    \textbf{b}, Test site and pedestrian-crossing configurations with and without occlusion.
    \textbf{c}, Predictive driver model score (PDMS) on 200 challenging NAVSIM scenes. 
    \textbf{d}, Real-world selection accuracy during the \emph{Normal} and \emph{Response} stages across six trials. Circles indicate individual trials and bars show the mean.
    \textbf{e}, Rewards for continued motion (solid) and braking (dashed) over ten decision times for NWM and Vista. The vertical line marks the onset of the dummy crossing.
    \textbf{f}, Imagined futures under the same continued-motion trajectory. Vista exhibits the \emph{same-speed motion} failure, whereas the NWM predicts the emerging conflict with the crossing dummy.}
    \label{fig:decision}
\end{figure}

This section examines whether more faithful imagination translates into better decisions. 
Following the procedure in Fig.~\ref{fig:overview}c, each world model imagines a future for each of two candidate ego trajectories from the same current observation. A common reward function accounting for safety, progress, and comfort is used to evaluate the resulting outcomes and select the higher-reward trajectory (Section~\ref{sec:selection}). 
As shown in Fig.~\ref{fig:decision}, we compare the NWM with Vista on the NAVSIM benchmark and real-world pedestrian-crossing experiments. No data from either setting were used to train or fine-tune the NWM.

We first evaluate this procedure on 200 challenging NAVSIM scenes in which ExploreVLA and AutoVLA each propose a candidate trajectory and only one candidate succeeds under the NAVSIM evaluator (see Fig.~\ref{fig:decision}c). 
We report the predictive driver model score (PDMS), the official NAVSIM metric for evaluating driving performance, where higher scores indicate better performance.
Using Vista-generated futures to evaluate the two candidates increases the PDMS to 0.637. NWM-guided decision-making further increases the PDMS to 0.783, substantially reducing the gap to the oracle score of 0.959. 
These results indicate that the more faithful futures generated by the NWM provide a more informative basis for decision-making when candidate driving policies disagree.

We further evaluate the same procedure on real sensor observations collected in controlled pedestrian-crossing experiments. The experimental platform consists of a Ford E-Transit electric research van equipped with a front-facing camera and onboard inertial sensing and computing hardware, together with a pedestrian dummy mounted on a wheeled cart (Fig.~\ref{fig:decision}a). The dummy crosses the vehicle path either unoccluded or from behind a parked car (Fig.~\ref{fig:decision}b). 
We conduct six trials under these crossing configurations (Supplementary Table 4). Each trial is divided into two decision stages. The \emph{Normal} stage contains five decision frames in which the ego vehicle remains relatively distant and the dummy has not begun to cross, whereas the \emph{Response} stage contains five frames in which the vehicle is closer and the dummy has begun to enter its path. 
The trajectory driven by the safety operator defines the correct choice, namely continued motion during the \emph{Normal} stage and braking during the \emph{Response} stage.
The selected decision frames and corresponding vehicle dynamics are shown in Supplementary Fig. 10.

Figure~\ref{fig:decision}d compares selection accuracy across these two stages. During the \textit{Normal} stage, Vista selects the correct action at all decision frames, and the NWM does so in 90\% of the cases. During the \textit{Response} stage, the NWM selects braking at all decision frames, while the accuracy of Vista decreases to 30\%. The degradation of Vista between the two stages indicates that it fails to reliably distinguish routine driving from an emerging pedestrian conflict. 
Consistent with the optimistic failure modes observed in Section~\ref{sec:survivorship_bias}, Vista often fails to imagine the conflict that would result from continued motion and therefore assigns insufficient penalty to that action. 
In contrast, the NWM achieves high accuracy in both stages, which indicates that its imagination is neither optimistic nor pessimistic.
All three errors made by the NWM occur at the final decision frame before the dummy
begins to cross. This frame is the closest to the transition between the \emph{Normal} and
\emph{Response} stages, where the visual evidence for the correct action is most ambiguous.
Importantly, these errors are safety-conservative, as the NWM selects braking earlier than necessary. By contrast, Vista instead selects continued motion toward a crossing pedestrian during the \emph{Response} stage, which is a more severe failure.

The temporal evolution of the rewards underlying these decisions further illustrates this difference (Fig.~\ref{fig:decision}e).
Before the crossing begins, both models assign a higher reward to continued motion than to braking.
Once the dummy begins to cross and the conflict develops, the NWM exhibits a clear reversal in action preference. The reward assigned to continued motion decreases sharply to near zero, while the reward assigned to braking progressively increases. 
This pattern is consistent across the six trials. 
In contrast, Vista exhibits a weaker separation between the two alternatives. After crossing onset, the reward assigned to continuing does not consistently fall below that assigned to braking, and the two overlap across the subsequent decision frames. Vista therefore does not produce a consistent preference for the appropriate braking action.

The imagined futures provide a qualitative explanation for these reward differences (Fig.~\ref{fig:decision}f). Given the same decision observation and the same continued-motion ego trajectory, Vista exhibits the \emph{same-speed motion} failure mode identified in Section~\ref{sec:survivorship_bias}. As the imagined sequence evolves, the pedestrian dummy remains at an approximately constant relative distance from the ego vehicle, which prevents the gap from closing and suppresses the developing conflict. The imagined outcome consequently remains benign despite continued ego motion. 
In contrast, the NWM imagines the vehicle approaching the dummy and reveals an emerging conflict with the crossing dummy. This difference directly affects the reward assigned to the continued-motion candidate. 
This optimistic bias can therefore mislead downstream decisions by making hazardous actions appear safe in the imagined future.
Across both NAVSIM and the real-world experiments, correcting survivorship bias enables the world model to provide a more reliable basis for safety-critical decision-making.

\section{Conclusion}

Our findings highlight a limitation of driving world models that is not captured by conventional assessments of video realism. A world model trained predominantly on routine driving can generate visually plausible futures while systematically replacing impending collisions with benign outcomes. Such errors change the consequences attributed to a candidate action, rather than merely affecting image appearance. By learning from real accident and routine-driving videos with an objective that emphasizes surrounding road users, the NWM can imagine safety-critical futures that are underrepresented in conventional driving datasets. These findings suggest that evaluating driving world models requires explicit assessment of safety-critical outcomes beyond conventional video quality metrics.

Importantly, learning to imagine dangerous outcomes should not come at the cost of treating routine driving as hazardous. A model that frequently imagines spurious collisions may favor unnecessary braking or excessively cautious maneuvers, just as a model that misses collisions may favor unsafe actions. Our evaluations suggest that the NWM better distinguishes between these situations while maintaining competitive generation quality. The goal is therefore not simply to increase the frequency of imagined collisions, but to generate futures that reflect the interactions and risks present in the observed scene.

The decision-making experiments illustrate the practical significance of faithful imagination. On challenging NAVSIM scenarios, evaluating candidate trajectories through imagined futures improves action selection relative to the baseline world model. The pedestrian-crossing experiments further demonstrate that this capability transfers to previously unseen real-world observations. In these scenarios, the NWM increasingly favors braking as the pedestrian conflict develops, whereas the baseline often fails to recognize the danger of continued motion. These findings suggest that improvements in the fidelity of imagined interactions can provide a more informative basis for selecting between candidate driving actions.
Beyond action selection, the ability to realize prescribed agent motions offers a way of constructing counterfactual variants of recorded scenes for safety testing.

One limitation of this work is the inference latency of the diffusion-based NWM, which currently requires $23.11 \pm 1.39$ s to generate a 25-frame video on a single GPU (Supplementary Table 2). This prevents real-time closed-loop deployment on a physical autonomous vehicle. Future work could address this through model distillation and accelerated sampling to enable efficient onboard inference.
Another limitation is that the NWM currently relies on a single front-facing camera and cannot anticipate conflicts that develop outside its field of view. Extending the framework to multi-view cameras and additional sensor modalities, such as LiDAR, could provide a more comprehensive understanding of complex driving interactions.

\bibliographystyle{unsrtnat}
\bibliography{references}

@article{ho2020denoising,
  title={Denoising diffusion probabilistic models},
  author={Ho, Jonathan and Jain, Ajay and Abbeel, Pieter},
  journal={Advances in neural information processing systems},
  volume={33},
  pages={6840--6851},
  year={2020}
}

@article{gosselin2025ctrl,
  title={Ctrl-crash: Controllable diffusion for realistic car crashes},
  author={Gosselin, Anthony and Luo, Ge Ya and Lara, Luis and Golemo, Florian and Nowrouzezahrai, Derek and Paull, Liam and Jolicoeur-Martineau, Alexia and Pal, Christopher},
  journal={arXiv preprint arXiv:2506.00227},
  year={2025}
}

@misc{zhang2025pure,
  title={Pure Vision Language Action (VLA) Models: A Comprehensive Survey}, 
  author={Dapeng Zhang and Jing Sun and Chenghui Hu and Xiaoyan Wu and Zhenlong Yuan and Rui Zhou and Fei Shen and Qingguo Zhou},
  year={2025},
  eprint={2509.19012},
  archivePrefix={arXiv},
  primaryClass={cs.RO}
}

@inproceedings{li2025causal,
  title={Causal-entity reflected egocentric traffic accident video synthesis},
  author={Li, Lei-Lei and Fang, Jianwu and Xiao, Junbin and Pang, Shanmin and Yu, Hongkai and Lv, Chen and Xue, Jianru and Chua, Tat-Seng},
  booktitle={2025 IEEE/CVF International Conference on Computer Vision (ICCV)},
  pages={11208--11218},
  year={2025},
  organization={IEEE}
}

@inproceedings{li2025avd2,
  title={Avd2: Accident video diffusion for accident video description},
  author={Li, Cheng and Zhou, Keyuan and Liu, Tong and Wang, Yu and Zhuang, Mingqiao and Gao, Huan-ang and Jin, Bu and Zhao, Hao},
  booktitle={2025 IEEE International Conference on Robotics and Automation (ICRA)},
  pages={13289--13296},
  year={2025},
  organization={IEEE}
}

@article{feng2021intelligent,
  title={Intelligent driving intelligence test for autonomous vehicles with naturalistic and adversarial environment},
  author={Feng, Shuo and Yan, Xintao and Sun, Haowei and Feng, Yiheng and Liu, Henry X},
  journal={Nature communications},
  volume={12},
  number={1},
  pages={748},
  year={2021},
  publisher={Nature Publishing Group UK London}
}

@article{feng2023dense,
  title={Dense reinforcement learning for safety validation of autonomous vehicles},
  author={Feng, Shuo and Sun, Haowei and Yan, Xintao and Zhu, Haojie and Zou, Zhengxia and Shen, Shengyin and Liu, Henry X},
  journal={Nature},
  volume={615},
  number={7953},
  pages={620--627},
  year={2023},
  publisher={Nature Publishing Group UK London}
}

@inproceedings{wang2024driving,
  title={Driving into the future: Multiview visual forecasting and planning with world model for autonomous driving},
  author={Wang, Yuqi and He, Jiawei and Fan, Lue and Li, Hongxin and Chen, Yuntao and Zhang, Zhaoxiang},
  booktitle={Proceedings of the IEEE/CVF Conference on Computer Vision and Pattern Recognition},
  pages={14749--14759},
  year={2024}
}

@article{zheng2026x,
  title={X-World: Controllable Ego-Centric Multi-Camera World Models for Scalable End-to-End Driving},
  author={Zheng, Chaoda and Li, Sean and Deng, Jinhao and Wang, Zhennan and Chen, Shijia and Xiao, Liqiang and Chi, Ziheng and Lin, Hongbin and Chen, Kangjie and Wang, Boyang and others},
  journal={arXiv preprint arXiv:2603.19979},
  year={2026}
}

@article{unterthiner2018towards,
  title={Towards accurate generative models of video: A new metric \& challenges},
  author={Unterthiner, Thomas and Van Steenkiste, Sjoerd and Kurach, Karol and Marinier, Raphael and Michalski, Marcin and Gelly, Sylvain},
  journal={arXiv preprint arXiv:1812.01717},
  year={2018}
}

@article{liu2024curse,
  title={Curse of rarity for autonomous vehicles},
  author={Liu, Henry X and Feng, Shuo},
  journal={nature communications},
  volume={15},
  number={1},
  pages={4808},
  year={2024},
  publisher={Nature Publishing Group UK London}
}

@inproceedings{yang2023unisim,
  title={Unisim: A neural closed-loop sensor simulator},
  author={Yang, Ze and Chen, Yun and Wang, Jingkang and Manivasagam, Sivabalan and Ma, Wei-Chiu and Yang, Anqi Joyce and Urtasun, Raquel},
  booktitle={2023 IEEE/CVF Conference on Computer Vision and Pattern Recognition (CVPR)},
  pages={1389--1399},
  year={2023},
  organization={IEEE}
}

@inproceedings{luo2025beyond,
  title={Beyond FVD: An enhanced evaluation metrics for video generation distribution quality},
  author={Luo, Ge Ya and Favero, Gian M and Luo, Zhi Hao and Jolicoeur-Martineau, Alexia and Pal, Christopher},
  booktitle={International Conference on Learning Representations},
  volume={2025},
  pages={66218--66252},
  year={2025}
}

@article{heusel2017gans,
  title={Gans trained by a two time-scale update rule converge to a local nash equilibrium},
  author={Heusel, Martin and Ramsauer, Hubert and Unterthiner, Thomas and Nessler, Bernhard and Hochreiter, Sepp},
  journal={Advances in neural information processing systems},
  volume={30},
  year={2017}
}

@article{khanam2024yolov11,
  title={Yolov11: An overview of the key architectural enhancements},
  author={Khanam, Rahima and Hussain, Muhammad},
  journal={arXiv preprint arXiv:2410.17725},
  year={2024}
}

@inproceedings{fang2024abductive,
  title={Abductive ego-view accident video understanding for safe driving perception},
  author={Fang, Jianwu and Li, Lei-lei and Zhou, Junfei and Xiao, Junbin and Yu, Hongkai and Lv, Chen and Xue, Jianru and Chua, Tat-Seng},
  booktitle={2024 IEEE/CVF Conference on Computer Vision and Pattern Recognition (CVPR)},
  pages={22030--22040},
  year={2024},
  organization={IEEE}
}

@article{tancik2020fourier,
  title={Fourier features let networks learn high frequency functions in low dimensional domains},
  author={Tancik, Matthew and Srinivasan, Pratul and Mildenhall, Ben and Fridovich-Keil, Sara and Raghavan, Nithin and Singhal, Utkarsh and Ramamoorthi, Ravi and Barron, Jonathan and Ng, Ren},
  journal={Advances in neural information processing systems},
  volume={33},
  pages={7537--7547},
  year={2020}
}

@inproceedings{kingma2013auto,
  title={Auto-encoding variational bayes},
  author={Kingma, Diederik P and Welling, Max},
  booktitle={International Conference on Learning Representations},
  year={2013}
}

@article{bardes2024revisiting,
  title={Revisiting feature prediction for learning visual representations from video},
  author={Bardes, Adrien and Garrido, Quentin and Ponce, Jean and Chen, Xinlei and Rabbat, Michael and LeCun, Yann and Assran, Mahmoud and Ballas, Nicolas},
  journal={arXiv preprint arXiv:2404.08471},
  year={2024}
}

@inproceedings{goyal2017something,
  title={The “something something” video database for learning and evaluating visual common sense},
  author={Goyal, Raghav and Kahou, Samira Ebrahimi and Michalski, Vincent and Materzynska, Joanna and Westphal, Susanne and Kim, Heuna and Haenel, Valentin and Fruend, Ingo and Yianilos, Peter and Mueller-Freitag, Moritz and others},
  booktitle={2017 IEEE International Conference on Computer Vision (ICCV)},
  pages={5843--5851},
  year={2017},
  organization={IEEE}
}

@inproceedings{radford2021learning,
  title={Learning transferable visual models from natural language supervision},
  author={Radford, Alec and Kim, Jong Wook and Hallacy, Chris and Ramesh, Aditya and Goh, Gabriel and Agarwal, Sandhini and Sastry, Girish and Askell, Amanda and Mishkin, Pamela and Clark, Jack and others},
  booktitle={International Conference on Machine Learning},
  pages={8748--8763},
  year={2021}
}

@inproceedings{carreira2017quo,
  title={Quo vadis, action recognition? a new model and the kinetics dataset},
  author={Carreira, Joao and Zisserman, Andrew},
  booktitle={2017 IEEE Conference on Computer Vision and Pattern Recognition (CVPR)},
  pages={4724--4733},
  year={2017},
  organization={IEEE}
}

@inproceedings{szegedy2016rethinking,
  title={Rethinking the inception architecture for computer vision},
  author={Szegedy, Christian and Vanhoucke, Vincent and Ioffe, Sergey and Shlens, Jon and Wojna, Zbigniew},
  booktitle={Proceedings of the IEEE Conference on Computer Vision and Pattern Recognition},
  pages={2818--2826},
  year={2016}
}

@inproceedings{veicht2024geocalib,
  title={Geocalib: Learning single-image calibration with geometric optimization},
  author={Veicht, Alexander and Sarlin, Paul-Edouard and Lindenberger, Philipp and Pollefeys, Marc},
  booktitle={European Conference on Computer Vision},
  pages={1--20},
  year={2024},
  organization={Springer}
}

@article{teed2021droid,
  title={Droid-slam: Deep visual slam for monocular, stereo, and rgb-d cameras},
  author={Teed, Zachary and Deng, Jia},
  journal={Advances in neural information processing systems},
  volume={34},
  pages={16558--16569},
  year={2021}
}

@article{yang2024depth,
  title={Depth anything v2},
  author={Yang, Lihe and Kang, Bingyi and Huang, Zilong and Zhao, Zhen and Xu, Xiaogang and Feng, Jiashi and Zhao, Hengshuang},
  journal={Advances in neural information processing systems},
  volume={37},
  pages={21875--21911},
  year={2024}
}

@inproceedings{zhang2023adding,
  title={Adding conditional control to text-to-image diffusion models},
  author={Zhang, Lvmin and Rao, Anyi and Agrawala, Maneesh},
  booktitle={Proceedings of the IEEE/CVF International Conference on Computer Vision},
  pages={3836--3847},
  year={2023}
}

@article{blattmann2023stable,
  title={Stable video diffusion: Scaling latent video diffusion models to large datasets},
  author={Blattmann, Andreas and Dockhorn, Tim and Kulal, Sumith and Mendelevitch, Daniel and Kilian, Maciej and Lorenz, Dominik and Levi, Yam and English, Zion and Voleti, Vikram and Letts, Adam and others},
  journal={arXiv preprint arXiv:2311.15127},
  year={2023}
}

@inproceedings{feng2022trafficgen,
  author={Feng, Lan and Li, Quanyi and Peng, Zhenghao and Tan, Shuhan and Zhou, Bolei},
  booktitle={2023 IEEE International Conference on Robotics and Automation (ICRA)}, 
  title={TrafficGen: Learning to Generate Diverse and Realistic Traffic Scenarios}, 
  year={2023},
  pages={3567-3575},
  doi={10.1109/ICRA48891.2023.10160296}}

@inproceedings{rempe2022generating,
  title={Generating useful accident-prone driving scenarios via a learned traffic prior},
  author={Rempe, Davis and Philion, Jonah and Guibas, Leonidas J and Fidler, Sanja and Litany, Or},
  booktitle={Proceedings of the IEEE/CVF Conference on Computer Vision and Pattern Recognition},
  pages={17305--17315},
  year={2022}
}

@article{chen2024end,
  title={End-to-end autonomous driving: Challenges and frontiers},
  author={Chen, Li and Wu, Penghao and Chitta, Kashyap and Jaeger, Bernhard and Geiger, Andreas and Li, Hongyang},
  journal={IEEE Transactions on Pattern Analysis and Machine Intelligence},
  volume={46},
  number={12},
  pages={10164--10183},
  year={2024},
  publisher={IEEE}
}

@article{ding2025understanding,
  title={Understanding world or predicting future? a comprehensive survey of world models},
  author={Ding, Jingtao and Zhang, Yunke and Shang, Yu and Zhang, Yuheng and Zong, Zefang and Feng, Jie and Yuan, Yuan and Su, Hongyuan and Li, Nian and Sukiennik, Nicholas and others},
  journal={ACM Computing Surveys},
  volume={58},
  number={3},
  pages={1--38},
  year={2025},
  publisher={ACM New York, NY}
}

@inproceedings{jiang2025survey,
  title={A survey on vision-language-action models for autonomous driving},
  author={Jiang, Sicong and Huang, Zilin and Qian, Kangan and Luo, Ziang and Zhu, Tianze and Zhong, Yang and Tang, Yihong and Kong, Menglin and Wang, Yunlong and Jiao, Siwen and others},
  booktitle={Proceedings of the IEEE/CVF International Conference on Computer Vision (ICCV) Workshops},
  pages={4524--4536},
  year={2025}
}

@article{zhou2025autovla,
  title={Autovla: A vision-language-action model for end-to-end autonomous driving with adaptive reasoning and reinforcement fine-tuning},
  author={Zhou, Zewei and Cai, Tianhui and Zhao, Seth Z and Zhang, Yun and Huang, Zhiyu and Zhou, Bolei and Ma, Jiaqi},
  journal={Advances in Neural Information Processing Systems},
  year={2025}
}

@inproceedings{hanselmann2022king,
  title={King: Generating safety-critical driving scenarios for robust imitation via kinematics gradients},
  author={Hanselmann, Niklas and Renz, Katrin and Chitta, Kashyap and Bhattacharyya, Apratim and Geiger, Andreas},
  booktitle={European Conference on Computer Vision},
  pages={335--352},
  year={2022},
  organization={Springer}
}

@inproceedings{wang2021advsim,
  title={Advsim: Generating safety-critical scenarios for self-driving vehicles},
  author={Wang, Jingkang and Pun, Ava and Tu, James and Manivasagam, Sivabalan and Sadat, Abbas and Casas, Sergio and Ren, Mengye and Urtasun, Raquel},
  booktitle={Proceedings of the IEEE/CVF Conference on Computer Vision and Pattern Recognition},
  pages={9909--9918},
  year={2021}
}

@article{ding2023survey,
  title={A survey on safety-critical driving scenario generation—a methodological perspective},
  author={Ding, Wenhao and Xu, Chejian and Arief, Mansur and Lin, Haohong and Li, Bo and Zhao, Ding},
  journal={IEEE Transactions on Intelligent Transportation Systems},
  volume={24},
  number={7},
  pages={6971--6988},
  year={2023},
  publisher={IEEE}
}

@article{xu2022safebench,
  title={Safebench: A benchmarking platform for safety evaluation of autonomous vehicles},
  author={Xu, Chejian and Ding, Wenhao and Lyu, Weijie and Liu, Zuxin and Wang, Shuai and He, Yihan and Hu, Hanjiang and Zhao, Ding and Li, Bo},
  journal={Advances in Neural Information Processing Systems},
  volume={35},
  pages={25667--25682},
  year={2022}
}

@inproceedings{hassan2025gem,
  title={Gem: A generalizable ego-vision multimodal world model for fine-grained ego-motion, object dynamics, and scene composition control},
  author={Hassan, Mariam and Stapf, Sebastian and Rahimi, Ahmad and Rezende, Pedro and Haghighi, Yasaman and Br{\"u}ggemann, David and Katircioglu, Isinsu and Zhang, Lin and Chen, Xiaoran and Saha, Suman and others},
  booktitle={Proceedings of the IEEE/CVF Conference on Computer Vision and Pattern Recognition},
  pages={22404--22415},
  year={2025}
}

@article{gao2024vista,
  title={Vista: A generalizable driving world model with high fidelity and versatile controllability},
  author={Gao, Shenyuan and Yang, Jiazhi and Chen, Li and Chitta, Kashyap and Qiu, Yihang and Geiger, Andreas and Zhang, Jun and Li, Hongyang},
  journal={Advances in Neural Information Processing Systems},
  volume={37},
  pages={91560--91596},
  year={2024}
}

@article{dauner2024navsim,
  title={Navsim: Data-driven non-reactive autonomous vehicle simulation and benchmarking},
  author={Dauner, Daniel and Hallgarten, Marcel and Li, Tianyu and Weng, Xinshuo and Huang, Zhiyu and Yang, Zetong and Li, Hongyang and Gilitschenski, Igor and Ivanovic, Boris and Pavone, Marco and others},
  journal={Advances in Neural Information Processing Systems},
  volume={37},
  pages={28706--28719},
  year={2024}
}

@inproceedings{yang2024generalized,
  title={Generalized predictive model for autonomous driving},
  author={Yang, Jiazhi and Gao, Shenyuan and Qiu, Yihang and Chen, Li and Li, Tianyu and Dai, Bo and Chitta, Kashyap and Wu, Penghao and Zeng, Jia and Luo, Ping and others},
  booktitle={Proceedings of the IEEE/CVF Conference on Computer Vision and Pattern Recognition},
  pages={14662--14672},
  year={2024}
}

@inproceedings{caesar2020nuscenes,
  title={nuscenes: A multimodal dataset for autonomous driving},
  author={Caesar, Holger and Bankiti, Varun and Lang, Alex H and Vora, Sourabh and Liong, Venice Erin and Xu, Qiang and Krishnan, Anush and Pan, Yu and Baldan, Giancarlo and Beijbom, Oscar},
  booktitle={Proceedings of the IEEE/CVF Conference on Computer Vision and Pattern Recognition},
  pages={11621--11631},
  year={2020}
}

@article{ha2018recurrent,
  title={Recurrent world models facilitate policy evolution},
  author={Ha, David and Schmidhuber, J{\"u}rgen},
  journal={Advances in neural information processing systems},
  volume={31},
  year={2018}
}

@article{ren2025cosmos,
  title={Cosmos-drive-dreams: Scalable synthetic driving data generation with world foundation models},
  author={Ren, Xuanchi and Lu, Yifan and Cao, Tianshi and Gao, Ruiyuan and Huang, Shengyu and Sabour, Amirmojtaba and Shen, Tianchang and Pfaff, Tobias and Wu, Jay Zhangjie and Chen, Runjian and others},
  journal={arXiv preprint arXiv:2506.09042},
  year={2025}
}

@online{waymo_world_model_2026,
  author  = {{Waymo}},
  title   = {The Waymo World Model: A New Frontier for Autonomous Driving Simulation},
  year    = {2026},
  url     = {https://waymo.com/blog/2026/02/the-waymo-world-model-a-new-frontier-for-autonomous-driving-simulation/},
  note    = {Accessed: 2026-04-29}
}

@inproceedings{sheng2026explorevla,
  title={ExploreVLA: Dense World Modeling and Exploration for End-to-End Autonomous Driving},
  author={Sheng, Zihao and Ye, Xin and Luo, Jingru and Chen, Sikai and Ren, Liu},
  booktitle={European Conference on Computer Vision (ECCV)},
  year={2026}
}

@article{hwang2024emma,
  title={Emma: End-to-end multimodal model for autonomous driving},
  author={Hwang, Jyh-Jing and Xu, Runsheng and Lin, Hubert and Hung, Wei-Chih and Ji, Jingwei and Choi, Kristy and Huang, Di and He, Tong and Covington, Paul and Sapp, Benjamin and others},
  journal={Waymo Technical Report},
  year={2024}
}

\clearpage
\appendix

\setcounter{figure}{0}
\renewcommand{\figurename}{Supplementary Fig.}
\renewcommand{\thefigure}{S\arabic{figure}}

\setcounter{table}{0}
\renewcommand{\tablename}{Supplementary Table}
\renewcommand{\thetable}{S\arabic{table}}

\begin{center}
  {\Large\bfseries Supplementary Information\par}
\end{center}


\section{Model implementation and training details}\label{smeth:training}

\subsection{Model components and conditioning implementation}
The neutral world model (NWM) is initialized from the pretrained checkpoint of Stable Video Diffusion Image-to-Video Model\footnote{\url{https://huggingface.co/stabilityai/stable-video-diffusion-img2vid-xt}}.
We use its temporal-decoder variational autoencoder (VAE), projected CLIP
image encoder and spatiotemporal U-Net. The VAE maps each
$512\times320$ RGB frame to a $4\times40\times64$ latent tensor; target
video latents are multiplied by the pretrained scaling factor $0.18215$.
The image encoder provides a 1,024-dimensional feature for cross-attention.
The ControlNet branch copies the corresponding input, downsampling and
middle-block weights of the adapted U-Net. Its four downsampling blocks
have channel widths $(320,640,1280,1280)$, and its residual output
projections are initialized to zero. These implementation details complement
the architecture in Fig.~1b and Methods Section~4.1.

For ego-motion conditioning, we translate the trajectory to its initial
position and rotate it into a local coordinate system with the initial
direction of travel along the positive forward axis. Four two-dimensional
positions at zero-based clip indices $5$, $10$, $15$ and $20$ are flattened
into an eight-dimensional vector. Coordinates remain in meters, without
additional distance scaling. Each scalar coordinate $a$ is encoded using
64 sinusoidal frequencies,
\begin{equation}
  \phi(a)=
  \bigl[(\cos(\omega_k a))_{k=0}^{63},
        (\sin(\omega_k a))_{k=0}^{63}\bigr],
  \qquad \omega_k=10000^{-k/64}.
  \label{seq:ego-fourier}
\end{equation}
The resulting 128-dimensional encodings are concatenated and passed through
a learned linear projection from 1,024 to 1,024 dimensions. This trajectory
feature is added to the image feature supplied to the ControlNet
cross-attention layers and repeated across all 25 frames.

Bounding-box controls are rasterized on a black RGB canvas at the source
image resolution. Each box has a category-specific outer color and a
filled interior with a track-specific color, which is reused for the same
identity across frames. 
The rendered images are resized to $512\times320$, normalized to $[-1,1]$
and encoded by the frozen VAE. A learned $4\times40\times64$ tensor,
initialized from a standard normal distribution, represents an unavailable
box condition.

During the second training stage, box controls are dropped as a contiguous
suffix. With probability $p(u)=\max(0.5,u)$, where $u\in[0,1]$ is the
fraction of the configured training schedule completed, a cutoff
$c$ is sampled uniformly from $\{0,\ldots,25\}$. Control latents at
indices $t\geq c$ are replaced by the learned null tensor. Otherwise, all
box controls are kept. The endpoints $c=0$ and $c=25$ correspond to
absent and complete box conditioning, respectively. Thus, the probability
of dropping the entire box sequence is $p(u)/26$, while intermediate cutoffs
provide partial conditioning. Image-conditioning dropout uses one uniform
draw per clip: both image branches are kept with probability $0.7$;
only the CLIP feature, only the initial-image VAE latent, or both are
dropped with probability $0.1$ each. In the second stage, the trajectory
feature is also omitted when the CLIP feature is dropped.

\subsection{Two-stage training procedure}
Both stages use the mixture of MM-AU and nuScenes described in Methods
Section~4.3.1, including the trajectory-based balancing of turning clips.
Stage~1 updates the full spatiotemporal U-Net while freezing the VAE and
image encoder. After 52,800 optimizer updates, this backbone initializes
Stage~2, in which the backbone remains frozen and the ControlNet, trajectory
projection and null embedding are optimized for 18,028 updates. The
reported model combines the Stage~1 backbone with the ControlNet from
\texttt{checkpoint-18028}. Training settings and computing resources are
summarized in Supplementary Table~\ref{stab:hyperparams}.

In both stages, a noise level is sampled uniformly from the 1,000 entries
of the pretrained training schedule. This schedule uses
$\beta_j=[\sqrt{0.00085}+(j/999)(\sqrt{0.012}-\sqrt{0.00085})]^2$,
$j=0,\ldots,999$, with
$\bar\alpha_j=\prod_{i=0}^{j}(1-\beta_i)$ and
$\sigma_j=\sqrt{(1-\bar\alpha_j)/\bar\alpha_j}$.
Gaussian noise is added as $z^{(s)}=z+\sigma_s\epsilon$, where
$\epsilon\sim\mathcal{N}(0,I)$. We use the pretrained $v$-prediction
parameterization and SVD preconditioning: the noisy latent input is scaled
by $(1+\sigma_s^2)^{-1/2}$, and the denoised estimate and loss weight are
\begin{equation}
  \hat z=
  \frac{z^{(s)}}{1+\sigma_s^2}
  -\frac{\sigma_s}{\sqrt{1+\sigma_s^2}}\,\hat v_\theta,
  \qquad
  \lambda(s)=\frac{1+\sigma_s^2}{\sigma_s^2}.
  \label{seq:diffusion-preconditioning}
\end{equation}
Here $\hat v_\theta$ denotes the denoising network output under the supplied
conditions.

Stage~1 adapts the pretrained SVD backbone to driving videos by
reconstructing clean video latents from noisy inputs, conditioned on the
initial observation. The denoising objective uses the noise-dependent
weighting $\lambda(s)$ defined above. The spatiotemporal U-Net is
fine-tuned while the VAE and CLIP image encoder remain frozen. 
Stage~2 uses the mask-based
$\mathcal{L}_{\mathrm{neutral}}$ defined in Methods Section~4.1.2, with
unit coefficient on the additional normalized box weights. Those weights
are detached from the computation graph and are set to zero at frames
whose box controls have been dropped, leaving the base reconstruction
weight of one at every latent location.

Optimization uses AdamW with constant learning rates, without warm-up or
learning-rate scaling by batch size. Each stage uses distributed data
parallelism across $8\times$ A100 GPUs, one clip per GPU and five gradient
accumulation steps, giving an effective batch size of 40 clips. Trainable
parameters are maintained in single precision with FP16 mixed-precision
computation and automatic loss scaling. Gradient checkpointing is enabled;
exponential moving averaging and explicit gradient-norm clipping are not
used. RGB preprocessing consists of resizing and normalization to
$[-1,1]$. The implementation uses Hugging Face Diffusers~0.27.2 and
Accelerate~0.30.0. The two stages take 134 and 53 hours, respectively,
corresponding to a total of 1,496 A100 GPU hours.

\begin{table}[htbp]
\centering
\small
\caption{\textbf{Training configuration of the neutral world model.}
Settings for backbone adaptation and conditional training; batch sizes are
measured in video clips. NWM, neutral world model; SVD, Stable Video
Diffusion; VAE, variational autoencoder; CLIP, Contrastive Language--Image
Pre-training; FP16, 16-bit floating point; EMA, exponential moving average;
GPU, graphics processing unit.}
\label{stab:hyperparams}
\begin{tabularx}{\linewidth}{@{}>{\raggedright\arraybackslash}p{0.27\linewidth}
  >{\raggedright\arraybackslash}X>{\raggedright\arraybackslash}X@{}}
\toprule
Setting & Stage 1: backbone adaptation & Stage 2: ControlNet \\
\midrule
Initialization & Pretrained SVD, 25-frame XT checkpoint & Stage 1 at step 52,800; copied ControlNet blocks and zero-initialized residual projections \\
Trainable modules & Full spatiotemporal U-Net & ControlNet, trajectory projection and null embedding \\
Frozen modules & VAE and CLIP image encoder & VAE, CLIP image encoder and adapted U-Net \\
Training data & MM-AU and nuScenes & MM-AU and nuScenes \\
Clip dimensions & 25 frames, $512\times320$ & 25 frames, $512\times320$ \\
Optimizer & AdamW & AdamW \\
AdamW parameters & $\beta_1=0.9$, $\beta_2=0.999$; $\epsilon=10^{-8}$; weight decay $0.01$ & $\beta_1=0.9$, $\beta_2=0.999$; $\epsilon=10^{-8}$; weight decay $0.01$ \\
Learning rate & $10^{-5}$, constant & $4\times10^{-5}$, constant \\
Batch size per GPU & 1 & 1 \\
Gradient accumulation & 5 steps & 5 steps \\
Effective batch size & 40 & 40 \\
Optimizer updates & 52,800 & 18,028 \\
Precision / EMA & FP16 mixed precision / no EMA & FP16 mixed precision / no EMA \\
Condition dropout & Coupled image-branch dropout & Image-branch dropout and stochastic box-suffix masking \\
Objective & Latent denoising loss & $\mathcal{L}_{\mathrm{neutral}}$ \\
Hardware & $8\times$ NVIDIA A100, 80 GB each & $8\times$ NVIDIA A100, 80 GB each \\
Training time & 134 h & 53 h \\
Accelerator time & 1,072 GPU h & 424 GPU h \\
\bottomrule
\end{tabularx}
\end{table}

\subsection{Inference, baseline configurations and computational cost}
NWM inference uses 30 Euler denoising steps with a Karras noise schedule
($\sigma_{\min}=0.002$, $\sigma_{\max}=700$, $\rho=7$; Supplementary
Table~\ref{stab:inference}). Guidance combines
the conditional NWM prediction with a reference prediction from the frozen,
original SVD U-Net. The reference model receives the same initial-image
conditioning but no trajectory or box controls. For frame index
$t\in\{0,\ldots,24\}$, the guided network output is
\begin{equation}
  v_t^{\mathrm{guided}}
  =v_t^{\mathrm{ref}}
   +g_t\bigl(v_t^{\mathrm{NWM}}-v_t^{\mathrm{ref}}\bigr),
  \qquad g_t=1+8t/24.
  \label{seq:reference-guidance}
\end{equation}
The ControlNet conditioning scale is one. The initial image receives
Gaussian noise augmentation with standard deviation $0.01$, and the motion
bucket identifier is 127. Inference uses FP16 weights with the pretrained
VAE's full-precision upcasting, U-Net forward chunking and VAE decoding in
chunks of eight frames. Each input produces one stochastic video sample.

Vista uses its released checkpoint and configs, with 50 Euler sampling steps,
$\sigma_{\min}=0.002$, $\sigma_{\max}=700$, $\rho=7$,
classifier-free guidance of 2.5 and conditioning-image noise augmentation
set to zero.\footnote{Vista weights:
\url{https://huggingface.co/OpenDriveLab/Vista}; sampling implementation:
\url{https://github.com/OpenDriveLab/Vista/blob/main/sample.py}.} GEM uses the released checkpoint and the official
Stage~2 configuration and sampling implementation.\footnote{GEM weights:
\url{https://huggingface.co/mariam-hassan/GEM}; sampling implementation:
\url{https://github.com/vita-epfl/GEM/blob/main/sample.py}.}
Its default dynamic-pyramid Euler sampler uses 50 steps, guidance of
1.5, $\sigma_{\max}=150$, $\rho=7$ and zero conditioning-image noise
augmentation. Both baselines receive the same source observation and ego
trajectory as NWM and generate 25 frames, with one sample per input.
Their native $1024\times576$ outputs are resized to $512\times320$ for
the common-resolution evaluations.

We measured NWM generation cost on a single NVIDIA RTX PRO 6000 Blackwell
Max-Q Workstation Edition GPU using PyTorch~2.8.0, CUDA~12.8,
Diffusers~0.27.2, Transformers~4.48.3 and Accelerate~0.30.0. After two
warm-up calls, five distinct inputs were processed sequentially
with batch size one. Wall-clock measurements were synchronized with
the GPU before and after each generation call. They include conditioning
encoding, diffusion sampling and VAE decoding, and exclude model loading,
disk input preparation, video encoding and file saving. A 25-frame video
took $23.11\pm1.39$~s (mean $\pm$ sample standard deviation, $n=5$), with
a maximum PyTorch-allocated GPU memory of 18.39~GiB. Peak reserved memory
was 33.06~GiB.

\begin{table}[htbp]
\centering
\small
\caption{\textbf{NWM inference settings and computational cost.}
Latency is measured after two warm-up calls over five distinct inputs on
the stated inference GPU. NWM, neutral world model; VAE, variational
autoencoder; RGB, red--green--blue; FP16, 16-bit floating point;
GPU, graphics processing unit; SD, sample standard deviation;
GiB, $2^{30}$ bytes.}
\label{stab:inference}
\begin{tabularx}{\linewidth}{@{}>{\raggedright\arraybackslash}p{0.35\linewidth}
  >{\raggedright\arraybackslash}X@{}}
\toprule
Setting & Value \\
\midrule
Sampler / denoising steps & Euler / 30 \\
Noise schedule & Karras; $\sigma_{\min}=0.002$, $\sigma_{\max}=700$, $\rho=7$ \\
Reference-model guidance & Linear increase from 1 to 9 across frames \\
ControlNet scale & 1.0 \\
Image noise augmentation & 0.01 \\
Motion bucket identifier & 127 \\
Output & 25 RGB frames, $512\times320$ \\
Samples per input & 1 \\
Precision / decoding & FP16 with VAE upcasting; eight-frame decoding chunks \\
Inference device & One NVIDIA RTX PRO 6000 Blackwell Max-Q Workstation Edition \\
Timing protocol & Batch size 1; two warm-up calls, five measured calls; initial-frame-only box conditioning \\
Latency per clip & $23.11\pm1.39$ s (mean $\pm$ SD) \\
Peak allocated GPU memory & 18.39 GiB \\
Peak reserved GPU memory & 33.06 GiB \\
\bottomrule
\end{tabularx}
\end{table}

\section{Datasets and pseudo-annotation}\label{smeth:datasets}

\subsection{Data sources}
Both training stages use nuScenes and MM-AU. We use the nuScenes
\texttt{v1.0-trainval} release and its official partition of 700 training
and 150 validation scenes.\footnote{\url{https://www.nuscenes.org/}}
Front-camera images, recorded ego poses, camera calibration and annotated
three-dimensional agent boxes provide the observations and conditioning
signals. For MM-AU\footnote{\url{https://huggingface.co/datasets/JeffreyChou/MM-AU}},
we use the videos, split manifest and agent annotations preprocessed by
Ctrl-Crash.\footnote{\url{https://github.com/AnthonyGosselin/Ctrl-Crash/blob/master/src/preprocess/README.md}}
The final clips used for NWM training and the statistics are listed in Supplementary Table~\ref{tab:data-statistics}.

\subsection{Filtering and clip extraction}
We follow the MM-AU quality exclusions provided by Ctrl-Crash and
construct clips from the remaining preprocessed image sequences.
After quality filtering and clip-eligibility checks, the final cache
contains 6,194 training videos and 733 validation videos
(Supplementary Table~\ref{tab:data-statistics}).
Each video contributes one 25-frame crash clip, and at most one nominal clip, selected as the earliest 25-frame window preceding the annotated abnormal interval and collision.
Post-collision windows are excluded from nominal-clip selection.

For nuScenes, the first 75 selected front-camera frames of each scene
provide 11 overlapping training windows with stride five, or three
non-overlapping validation windows. This yields 7,700 training and 450
validation clips. Ego trajectories are obtained from the recorded poses.
Agent boxes are obtained by projecting the annotated three-dimensional
boxes into the calibrated front camera, clipping their projected extent to
the image, and preserving instance identities across frames. All training
clips contain 25 RGB frames resized to $512\times320$ pixels and normalized
to $[-1,1]$.

\begin{table}[htbp]
  \centering
  \caption{\textbf{Dataset splits and training composition.}
  Counts of extracted 25-frame clips and effective training samples
  after maneuver-based repetition.}
  \label{tab:data-statistics}
  \small
  \begin{tabularx}{\linewidth}{@{}Xrrr@{}}
    \toprule
    \textbf{Dataset} & Training & Validation & Effective training \\
    \midrule
    nuScenes & 7,700 & 450 & 11,500 \\
    MM-AU (crash) & 6,194 & 733 & 9,630 \\
    MM-AU (nominal) & 1,745 & 219 & 2,381 \\
    MM-AU (total) & 7,939 & 952 & 12,011 \\
    \midrule
    Combined total & 15,639 & 1,402 & 23,511 \\
    \bottomrule
  \end{tabularx}
\end{table}

\subsection{Maneuver assignment and sampling balance}
Maneuvers are assigned from the local ego-trajectory condition. After
translating the trajectory to its initial position and aligning its initial
direction with the positive forward axis, let $(x_{20},y_{20})$ be the
position at zero-based clip index 20, the last of the four conditioning
points. Here, $x$ denotes lateral displacement, positive to the right, and
$y$ denotes forward displacement, in meters. The assignment is
\begin{equation}
  c=\begin{cases}
    \text{right turn}, & x_{20}\geq 2,\\
    \text{left turn}, & x_{20}\leq -2,\\
    \text{stopping}, & -2<x_{20}<2\ \text{and}\ y_{20}\leq 2,\\
    \text{straight}, & -2<x_{20}<2\ \text{and}\ y_{20}>2.
  \end{cases}
  \label{seq:maneuver-assignment}
\end{equation}
The stopping category therefore denotes small forward displacement under
this rule. Turning conditions take precedence over the forward-displacement criterion.

Before balancing, the combined training set contains 1,095 right-turn,
873 left-turn, 5,037 stopping and 8,634 straight-driving clips. Turning
clips account for only 12.6\% of this set. For both datasets, left- and right-turn clips are oversampled by a factor of five during training, while all other clips are included once. 
The resulting effective counts are 5,475, 4,365,
5,037 and 8,634, respectively, giving proportions of 23.3\%, 18.6\%,
21.4\% and 36.7\%. The effective dataset contains 11,500 nuScenes samples
(48.9\%) and 12,011 MM-AU samples (51.1\%). Across this combined mixture,
9,630 effective samples are crash clips and 13,881 are nominal clips
(41.0\% and 59.0\%, respectively). 
The dataset source, crash-versus-nominal and maneuver distributions before and after balancing are summarized in Supplementary Fig.~\ref{fig:data-composition}.

\begin{figure}[htbp]
  \centering
  \includegraphics[width=0.99\linewidth]{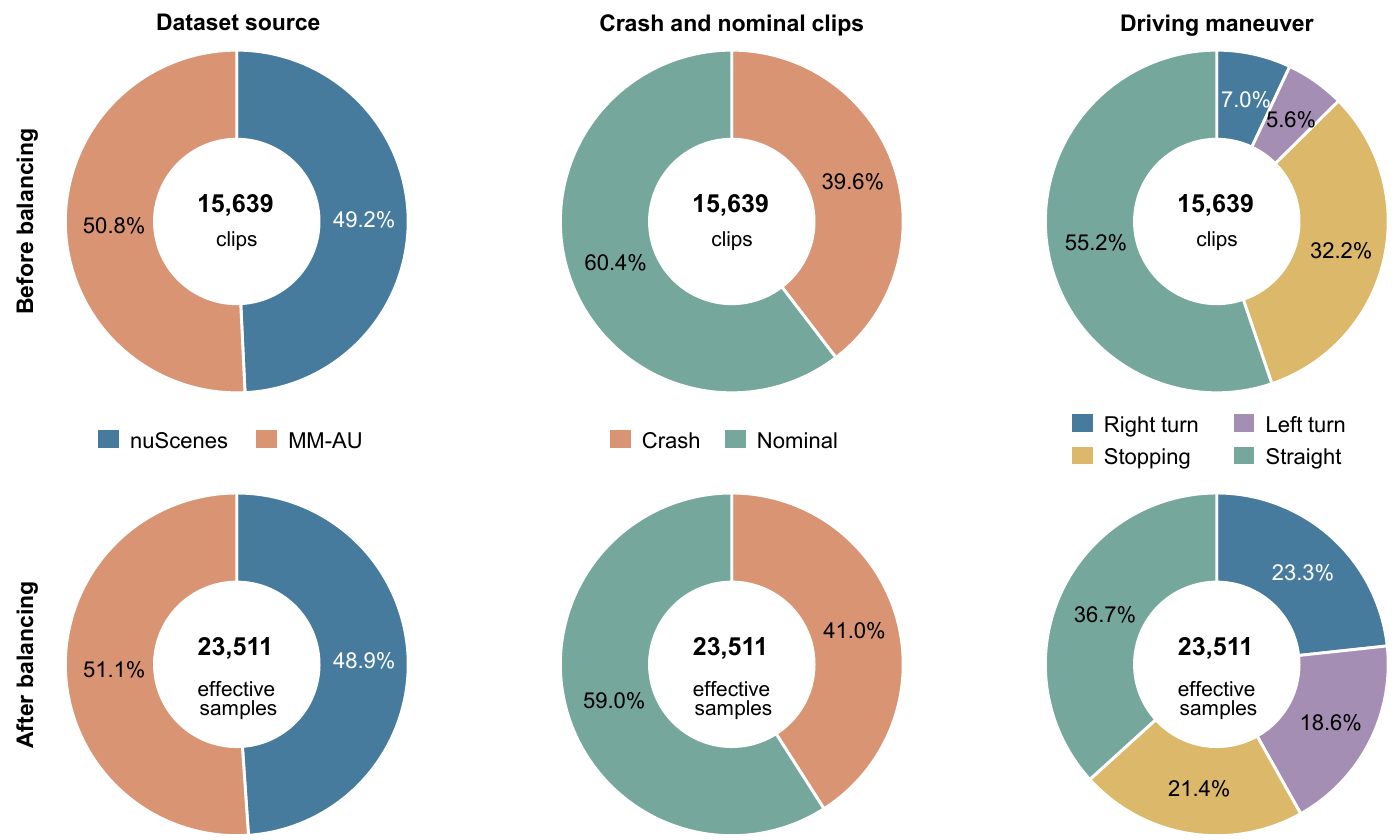}
  \caption{\textbf{Composition and balancing of training data.}}
  \label{fig:data-composition}
\end{figure}

\subsection{Ego-trajectory pseudo-annotation for MM-AU}
We estimate MM-AU ego motion using GEM's released pseudo-labeling
pipeline, which combines GeoCalib camera calibration~\cite{veicht2024geocalib}, Depth Anything V2
metric depth prediction~\cite{yang2024depth} and DROID-SLAM trajectory estimation~\cite{teed2021droid}.\footnote{%
\url{https://github.com/vita-epfl/GEM/blob/main/pseudo_labeling_pipeline/README.md}}
The input consists of preprocessed RGB sequences. GeoCalib uses its pretrained pinhole model and
100 temporally distributed frame samples per video, with shared intrinsics,
to estimate one camera matrix. The pipeline averages the returned camera
matrices over these samples and rescales the intrinsics when resizing
images for trajectory estimation.

Depth is predicted with the outdoor ViT-S checkpoint
\nolinkurl{depth_anything_v2_metric_vkitti_vits.pth}, trained for metric
depth estimation with a maximum depth of 80\,m. The depth maps and estimated
intrinsics are supplied with the RGB sequence to DROID-SLAM, initialized
from \texttt{droid.pth}. The trajectory stage uses a maximum segment length
of 2,000 frames, an overlap of 100 frames between segments and a minimum
segment-length setting of 100 frames. DROID-SLAM performs global bundle
adjustment at termination and exports camera poses as $4\times4$
transformation matrices with frame indices. Metric scale follows the
learned depth predictions. These trajectories are visual estimates, in
contrast to the sensor-recorded ego poses supplied by nuScenes.

For conditioning, we extract the horizontal and forward translation
components from the estimated camera trajectory, corresponding to its
$x$ and $z$ axes. Before clip extraction and frame subsampling, both
components are smoothed along the full preprocessed sequence with a
one-dimensional Gaussian filter of standard deviation 10 frames and
nearest-value boundary extension. Each selected clip is then translated
to start at the origin and rotated to align its initial displacement
with the positive forward axis. This normalization changes the coordinate
origin and orientation while preserving distance scale. Positions at
zero-based indices $5$, $10$, $15$ and $20$ form the eight-dimensional
trajectory condition used by NWM. Supplementary Fig.~\ref{fig:pseudo-annotations}
illustrates the RGB input, estimated depth and resulting trajectory.

\subsection{Agent bounding-box pseudo-annotation for MM-AU}
We directly reuse the frame-wise agent annotations provided by Ctrl-Crash,
including class names, normalized two-dimensional box coordinates and
video-local track identities. The selected classes are person, car, truck,
bus, train, motorcycle and bicycle. Annotations are matched to the selected
frames by image filename, converted to image coordinates and rendered as
control frames, using category colors for box borders and consistent
track colors for their interiors. Only boxes present in the
frame annotation are rendered. These boxes provide both the visual
conditioning input and the spatial masks for the bounding-box-guided loss.
The first row of each example in Supplementary Fig.~\ref{fig:pseudo-annotations}
shows rendered boxes overlaid on the original RGB frames.

\begin{figure}[htbp]
  \centering
  \includegraphics[width=0.99\linewidth]{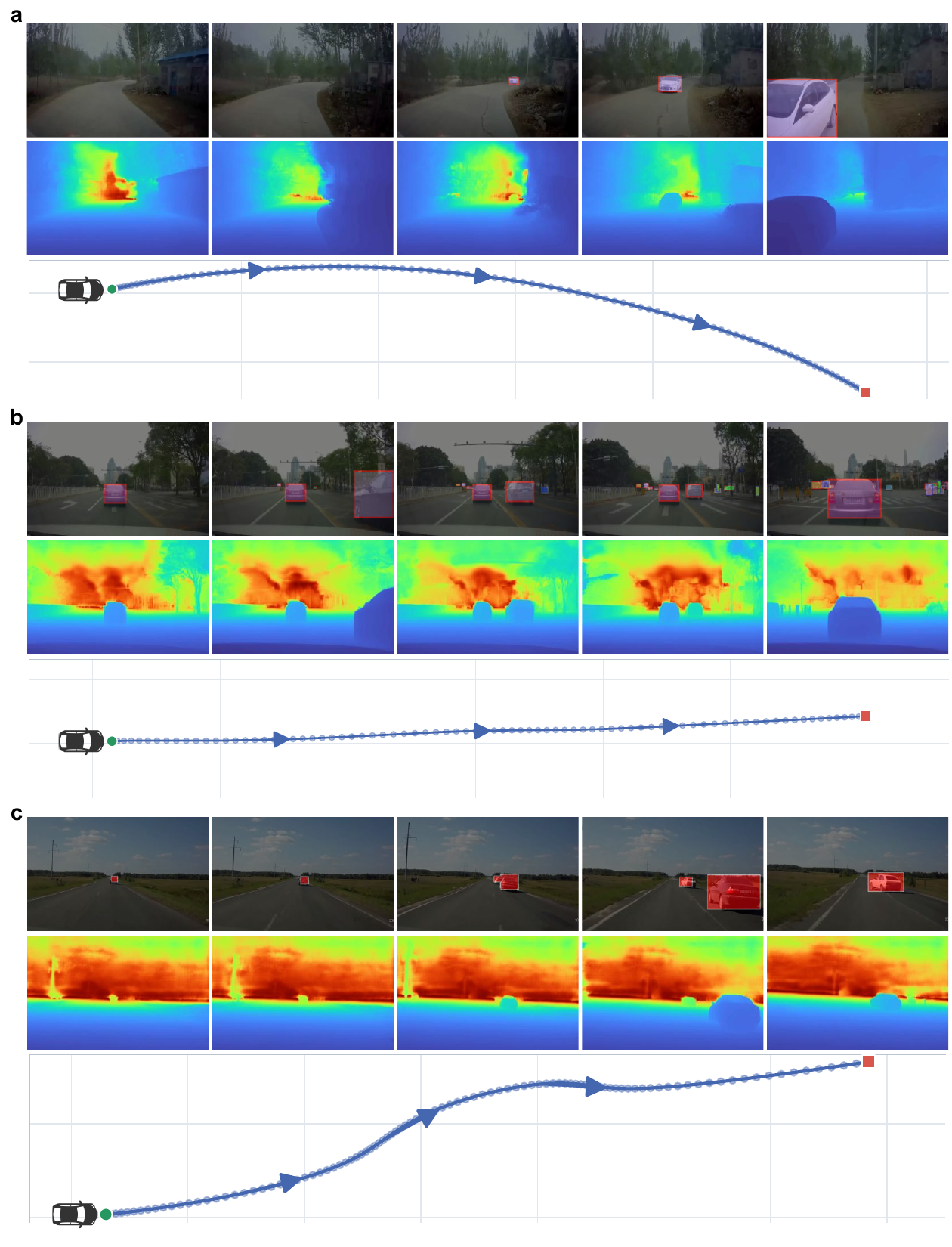}
  \caption{\textbf{Examples of pseudo-annotation of MM-AU videos.}
    \textbf{a}--\textbf{c}, Three representative clips in which the ego vehicle turns right (\textbf{a}), drives straight (\textbf{b}) and turns left (\textbf{c}).
    For each clip, the top row shows the original RGB observations overlaid with the detected bounding boxes, the middle row shows the metric depth estimated with Depth Anything V2 (warmer colors indicate greater distance), and the bottom row shows the ego trajectory estimated using calibrated, depth-assisted DROID-SLAM in the bird's-eye view.
    Frames are ordered chronologically from left to right.
    In the trajectory plots, the green circle and red square mark the start and end positions.}
  \label{fig:pseudo-annotations}
\end{figure}

\section{Evaluation protocols and statistical analysis}\label{smeth:evaluation}

\subsection{Accident-scene sampling and prediction start points}
The accident evaluation uses 100 ego-involved MM-AU videos selected from
the validation set. Ego involvement is determined by the
\texttt{ego\_involved} annotation, and the collision time is obtained from
the accident-frame metadata. Candidates must have a valid collision
annotation, complete image sequences and conditioning information for all four prediction starting points.

\begin{figure}[htbp]
  \centering
  \includegraphics[width=0.99\linewidth]{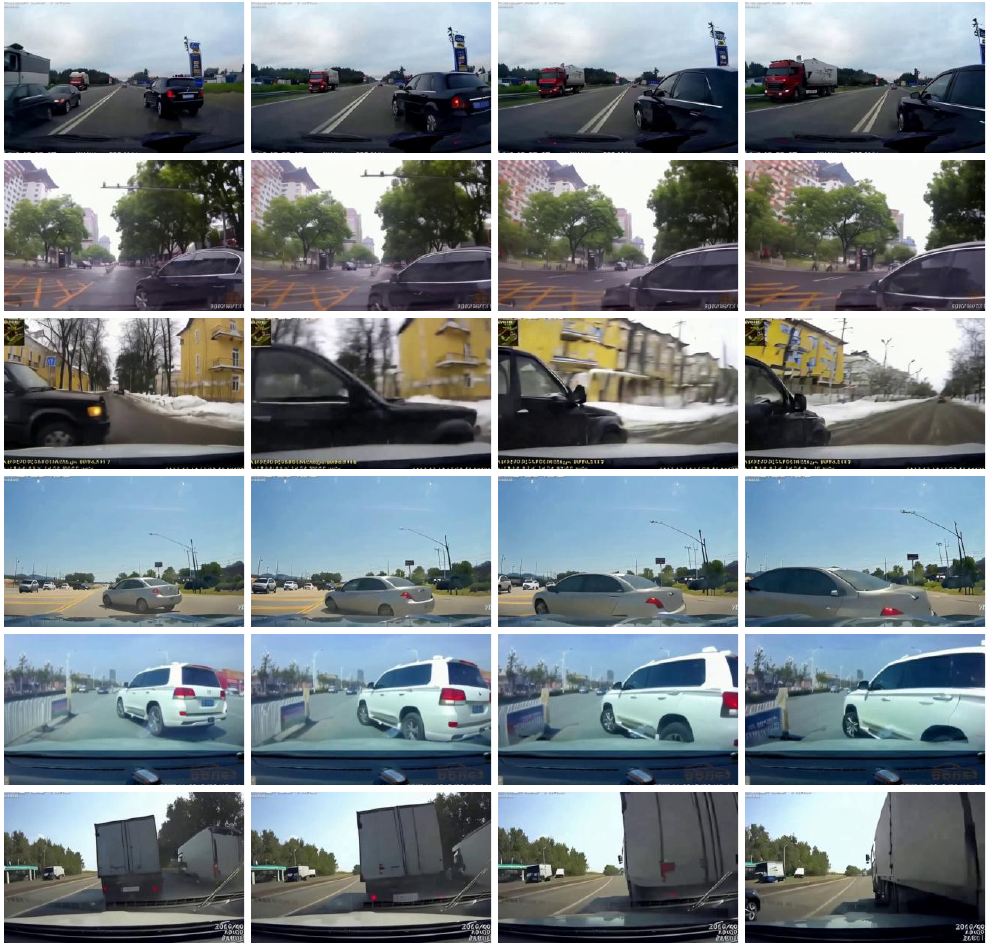}
  \caption{\textbf{Additional accident-driving generations.}
    Each row shows one clip generated by the NWM, with frames ordered chronologically from left to right. }
  \label{sfig:crash-qualitative}
\end{figure}

For each video, prediction starts approximately $h\in\{2.0,1.5,1.0,0.5\}$\,s
before the annotated collision. The requested start is snapped to the
nearest available frame, with ties resolved toward the later frame; the
maximum rounding error is $1/12$\,s. Each window contains 25 frames. The
initial RGB observation and ego-trajectory condition are constructed
separately for each start point. In particular, the trajectory is extracted
from the full-video estimate, smoothed, translated and rotated relative to
the new start, and sampled at clip indices $5$, $10$, $15$ and $20$.
Each model receives the matched observation and ego trajectory and generates one video per input. 
This gives 400 generated clips per model, with 100 source videos shared
across the four prediction-start groups.
Supplementary Fig.~\ref{sfig:crash-qualitative}
provides additional visualization examples of generated videos.

\subsection{Routine-driving evaluation and video-quality metrics}
The routine-driving evaluation uses 50 MM-AU and 100 nuScenes validation
clips. The sample list allocation reflects the relative sizes of the available validation
pools. MM-AU candidates are nominal clips whose annotated collision
frame lies outside the selected window. nuScenes candidates are
non-overlapping 25-frame front-camera windows from validation scenes.
All three models are evaluated on the same final clip lists
and their paired reference videos, using one generated video per input.
The false collision rate is
\begin{equation}
  \operatorname{FCR}_{m,d}
  =\frac{1}{N_d}\sum_{i=1}^{N_d}C_{m,d,i},
  \qquad N_{\mathrm{MM\text{-}AU}}=50,\quad
  N_{\mathrm{nuScenes}}=100,
  \label{seq:false-collision-rate}
\end{equation}
where $d$ denotes the dataset and $C_{m,d,i}$ is the manual collision label
for a nominal input. 

For video-quality evaluation, generated and reference videos are decoded
to RGB, and 25 frames are selected at uniformly spaced indices spanning
each clip. Every selected frame is resized to $512\times320$
using bilinear interpolation (OpenCV \texttt{INTER\_LINEAR}).
Scores are computed from the generated and reference feature distributions
over the full selected dataset.

Fr\'echet Inception Distance (FID) uses the 2,048-dimensional Inception-v3
features~\cite{szegedy2016rethinking} through the
\texttt{torch-fidelity} implementation. All 25 frames contribute,
giving 1,250 frames per distribution for MM-AU and 2,500 for nuScenes.
Fr\'echet Video Distance (FVD) uses the 400-dimensional I3D outputs~\cite{carreira2017quo} from
\texttt{i3d\_torchscript.pt}. Input videos are normalized to $[-1,1]$,
resized with aspect ratio preserved so that the shorter side is 224 pixels,
and center-cropped to $224\times224$. One feature vector is extracted per
video. Both distances compare feature means and covariances:
\begin{equation}
  D_{\mathrm{F}}
  =\lVert\mu_{\mathrm{r}}-\mu_{\mathrm{g}}\rVert_2^2
   +\operatorname{tr}\!\left[
      \Sigma_{\mathrm{r}}+\Sigma_{\mathrm{g}}
      -2\left(\Sigma_{\mathrm{r}}^{1/2}\Sigma_{\mathrm{g}}
                 \Sigma_{\mathrm{r}}^{1/2}\right)^{1/2}\right],
  \label{seq:frechet-video-quality}
\end{equation}
where subscripts $\mathrm{r}$ and $\mathrm{g}$ denote reference and
generated features, respectively.

\begin{figure}[htbp]
  \centering
  \includegraphics[width=0.99\linewidth]{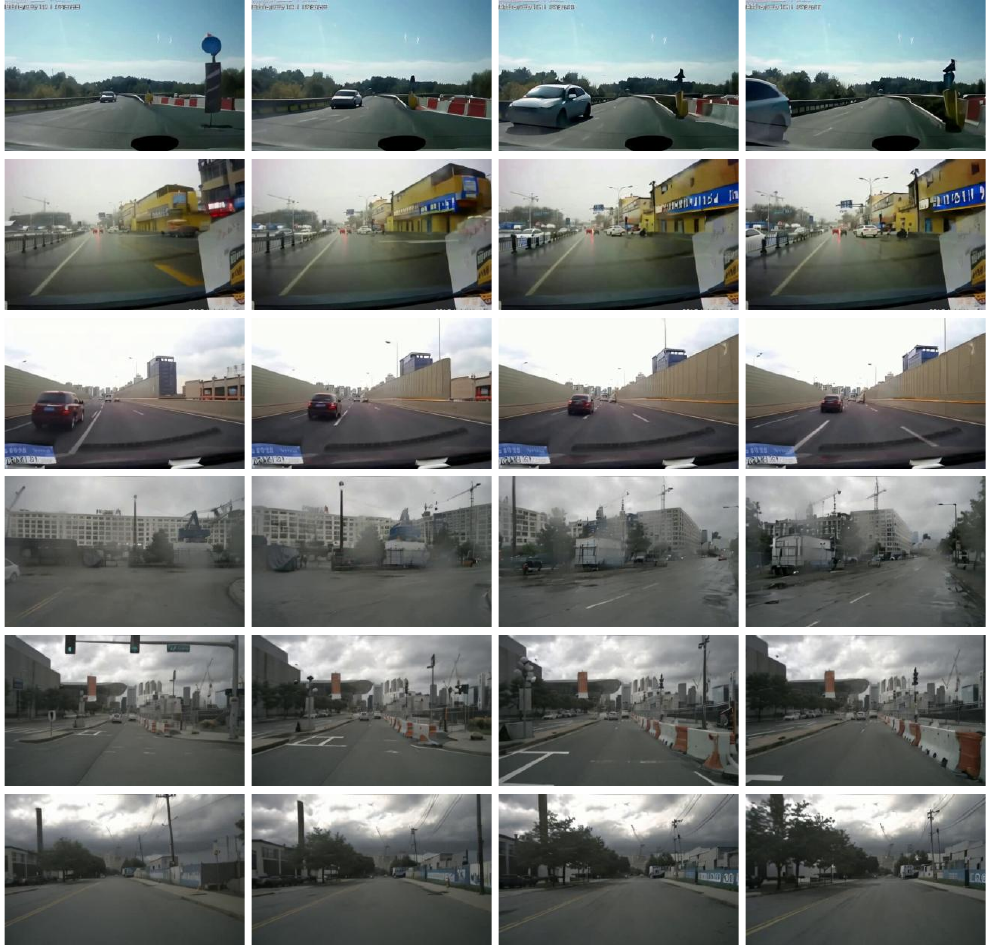}
  \caption{\textbf{Additional routine-driving generations.}
    Each row shows one clip generated by the NWM, with frames ordered chronologically from left to right. }
  \label{sfig:routine-qualitative}
\end{figure}

JEDi is computed with \texttt{videojedi} 1.1.0 and \texttt{vjepa} 0.1.2,
using the V-JEPA ViT-H/16 encoder~\cite{bardes2024revisiting} and its Something-Something V2~\cite{goyal2017something} probe
pooler. The weights are \texttt{vith16.pth.tar} and
\texttt{ssv2-probe.pth.tar}; each video yields a 1,280-dimensional feature.
The 25-frame input is resized and center-cropped to $224\times224$,
scaled to $[0,1]$, and normalized with RGB means
$(0.485,0.456,0.406)$ and standard deviations $(0.229,0.224,0.225)$.
For reference features $u_i$ and generated features $v_j$, the implementation
uses
\begin{equation}
  \begin{aligned}
  \operatorname{JEDi}&=100\left[
    \frac{1}{n^2}\sum_{i,i'=1}^{n}k(u_i,u_{i'})
    +\frac{1}{m^2}\sum_{j,j'=1}^{m}k(v_j,v_{j'})
    -\frac{2}{nm}\sum_{i=1}^{n}\sum_{j=1}^{m}k(u_i,v_j)
  \right],\\
  k(u,v)&=\left(\frac{u^{\mathsf T}v}{1280}\right)^2.
  \end{aligned}
  \label{seq:jedi}
\end{equation}
Here $n=m=50$ for MM-AU and $n=m=100$ for nuScenes. Lower FID, FVD and JEDi indicate closer
agreement between the generated and reference distributions.
Supplementary Fig.~\ref{sfig:routine-qualitative} provides additional visualization examples of generated videos.

\subsection{Bounding-box consistency under agent conditioning}
We compare NWM generations using null box condition with
generations using the complete 25-frame box sequence. Both settings use
the same 50 MM-AU and 100 nuScenes inputs, ego trajectories and generation
seeds. Spatial consistency is assessed between detected boxes
in corresponding reference and generated RGB frames, as illustrated in
Supplementary Fig.~\ref{sfig:box-metric-protocol}.

\begin{figure}[htbp]
  \centering
  \includegraphics[width=0.99\linewidth]{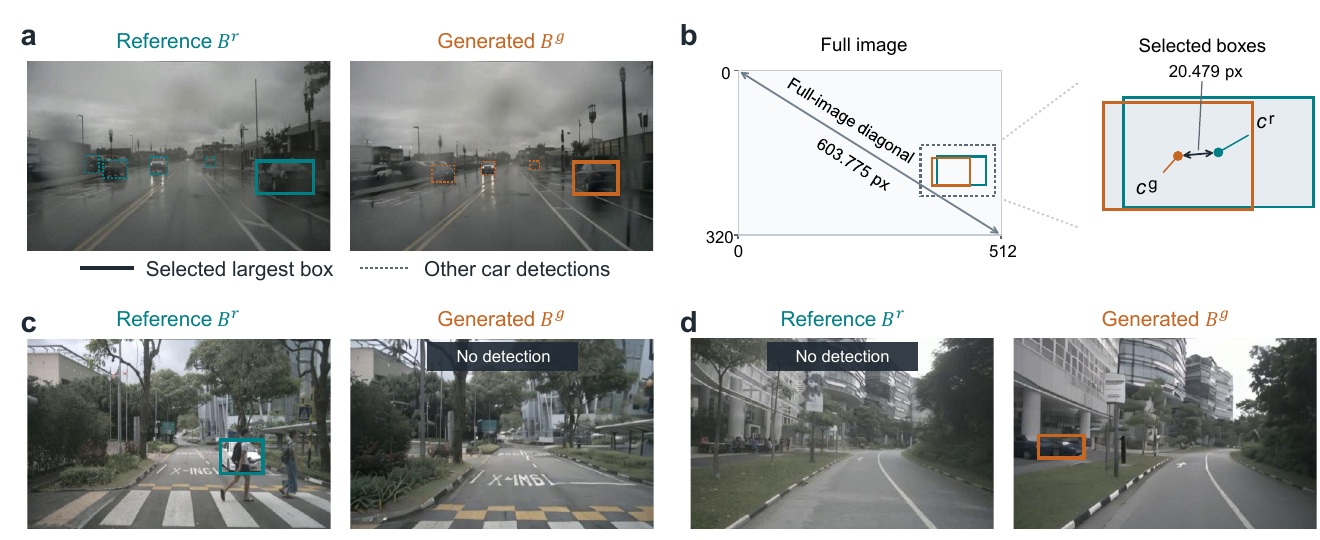}
  \caption{\textbf{Bounding-box selection and spatial-consistency metrics.}
  \textbf{a}, The largest detection is selected independently in the
  reference and generated frame. \textbf{b}, Selected boxes are compared
  in a common image coordinate system using intersection over union (IoU),
  success at IoU $\geq0.5$, and center distance normalized by the full image
  diagonal. \textbf{c}, A missing generated detection receives IoU $=0$,
  success $=0$ and normalized center offset (nCO) $=1$.
  \textbf{d}, A frame without a reference detection is excluded.}
  \label{sfig:box-metric-protocol}
\end{figure}

Detection uses \texttt{yolo11x.pt} with Ultralytics 8.4.19, confidence
threshold 0.25, non-maximum-suppression IoU threshold 0.7, inference image
size 640 and a maximum of 100 detections per frame. For this comparison,
we independently select the detection with the largest area in each frame. Let $B^{\mathrm{r}}_t$ and
$B^{\mathrm{g}}_t$ denote these reference and generated boxes, with centers
$c^{\mathrm{r}}_t$ and $c^{\mathrm{g}}_t$. Selection is performed frame by
frame. Box coordinates are expressed in the common $512\times320$ image coordinate system.

A frame is eligible when a reference detection is available. When both
boxes are detected, its intersection over union and normalized center
offset are
\begin{equation}
  I_t=\frac{|B^{\mathrm{r}}_t\cap B^{\mathrm{g}}_t|}
             {|B^{\mathrm{r}}_t\cup B^{\mathrm{g}}_t|},
  \qquad
  d_t=\frac{\lVert c^{\mathrm{g}}_t-c^{\mathrm{r}}_t\rVert_2}
                 {\sqrt{512^2+320^2}}.
  \label{seq:box-frame-metrics}
\end{equation}
If the generated frame has no detection, we assign $I_t=0$ and
$d_t=1$. Frames with no reference detection are excluded. Over the eligible
frame set $\mathcal E$, pooled across clips within each dataset and
conditioning setting, we report mean IoU (mIoU), success rate at an IoU
threshold of 0.5 (SR@0.5), and mean normalized center offset (nCO):
\begin{equation}
  \operatorname{mIoU}=\frac{1}{|\mathcal E|}\sum_{t\in\mathcal E}I_t,
  \qquad
  \text{SR@0.5}=\frac{1}{|\mathcal E|}
       \sum_{t\in\mathcal E}\mathbf{1}[I_t\geq0.5],
  \qquad
  \operatorname{nCO}=\frac{1}{|\mathcal E|}\sum_{t\in\mathcal E}d_t.
  \label{seq:box-pooled-metrics}
\end{equation}
Higher mIoU and SR@0.5 and lower nCO indicate greater spatial consistency.

\begin{figure}[htbp]
  \centering
  \includegraphics[width=0.99\linewidth]{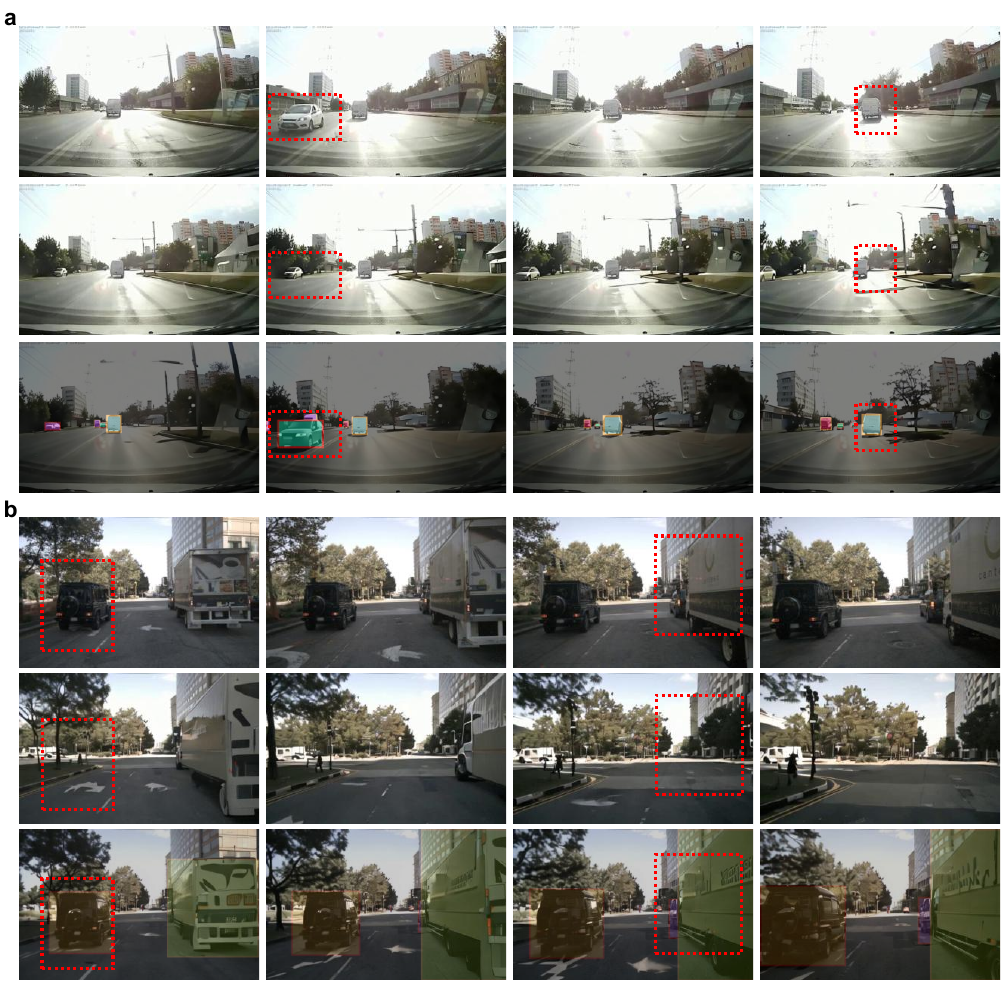}
  \caption{\textbf{Additional examples of bounding-box conditioning.}
    \textbf{a}, Clips from MM-AU. \textbf{b}, Clips from nuScenes.
    For each clip, the top row shows the ground-truth frames, the middle row shows the frames generated by the NWM without the bounding-box condition, and the bottom row shows the frames generated with the full sequence of bounding-box conditions, overlaid with the bounding boxes.
    Red dashed rectangles mark the same regions across rows, highlighting agents that are absent or misplaced without the bounding-box conditions but correctly generated when bounding-box conditions are provided.}
  \label{sfig:control-qualitative}
\end{figure}

\begin{figure}[htbp]
  \centering
  \includegraphics[width=0.99\linewidth]{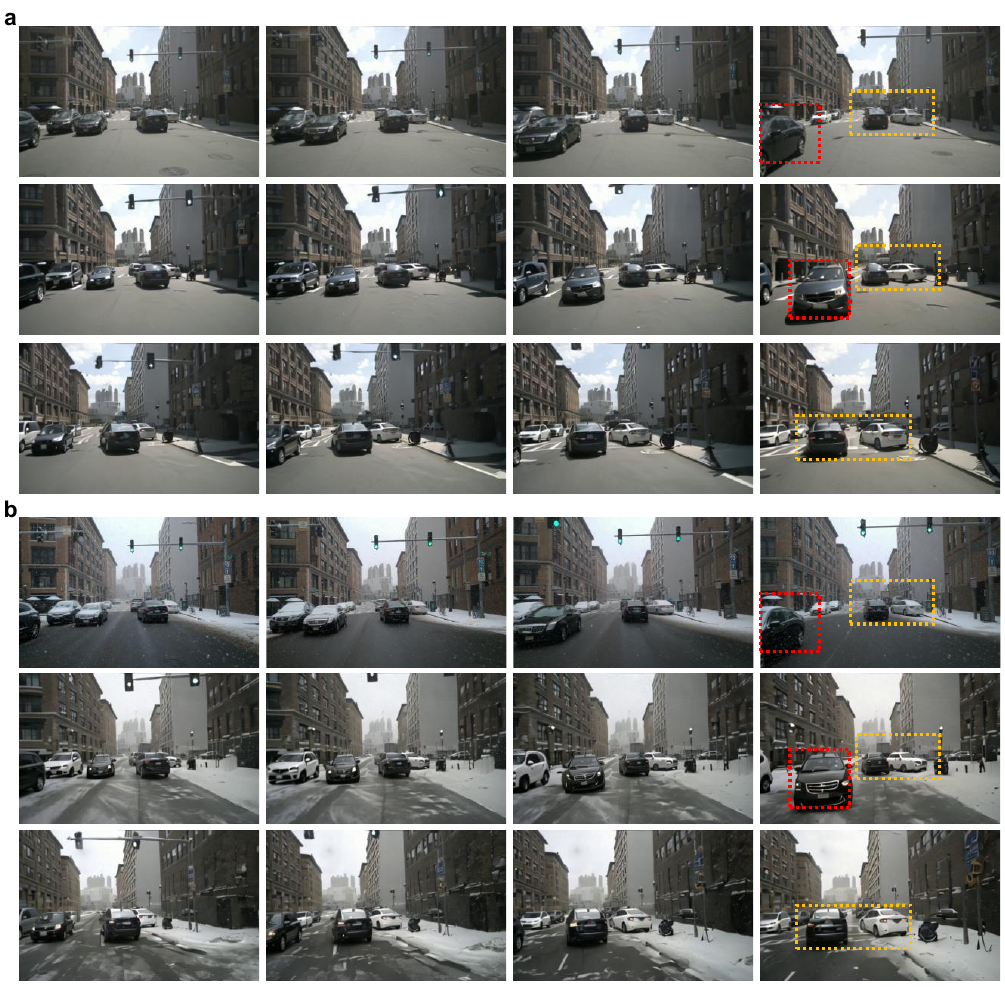}
    \caption{\textbf{Additional examples of STRIVE-conditioned generation.}
    \textbf{a}, A nuScenes scene. \textbf{b}, The same scene generated under snowy conditions.
    Top, original motion, shown as the recorded nuScenes frames (\textbf{a}) or
    their snow-edited counterparts (\textbf{b}); middle and bottom, adversarial
    and solution motions generated by the NWM using STRIVE-generated agent
    trajectories as the bounding-box condition.
    Red dashed rectangles mark the adversarial vehicle, which cuts into the ego
    path in the adversarial motion. Yellow dashed rectangles mark the leading
    vehicles, whose displacement in the solution motion reflects the evasive
    maneuver of the ego vehicle.}
  \label{sfig:strive-qualitative}
\end{figure}

Supplementary Fig.~\ref{sfig:control-qualitative} shows additional
qualitative examples from both datasets. Without the box condition, agents
can be absent or misplaced relative to the ground truth, whereas
conditioning on the full box sequence recovers their presence and position.
Beyond boxes taken from recorded data, Supplementary
Fig.~\ref{sfig:strive-qualitative} shows generations conditioned on
STRIVE-generated agent trajectories. The original motion is the recorded
nuScenes clip, and the NWM generates the adversarial and solution motions.
The adversarial vehicle cuts into the ego path, and the displacement of the
leading vehicles in the solution motion reflects the evasive maneuver of
the ego vehicle. 
We further show a synthetic snowy variant of the same scene in Supplementary
Fig.~\ref{sfig:strive-qualitative}b. The recorded
frames are edited into snowy weather with Qwen-Image-Edit-2511\footnote{\url{https://huggingface.co/Qwen/Qwen-Image-Edit-2511}}, and the first
edited frame serves as the initial observation of the NWM, which is
conditioned on the same STRIVE-generated trajectories. This example is
provided for qualitative illustration only and is not included in the
quantitative evaluation.

\subsection{Manual assessment of collision outcomes}\label{smeth:annotation}
\begin{figure}[htbp]
  \centering
  \includegraphics[width=0.99\linewidth]{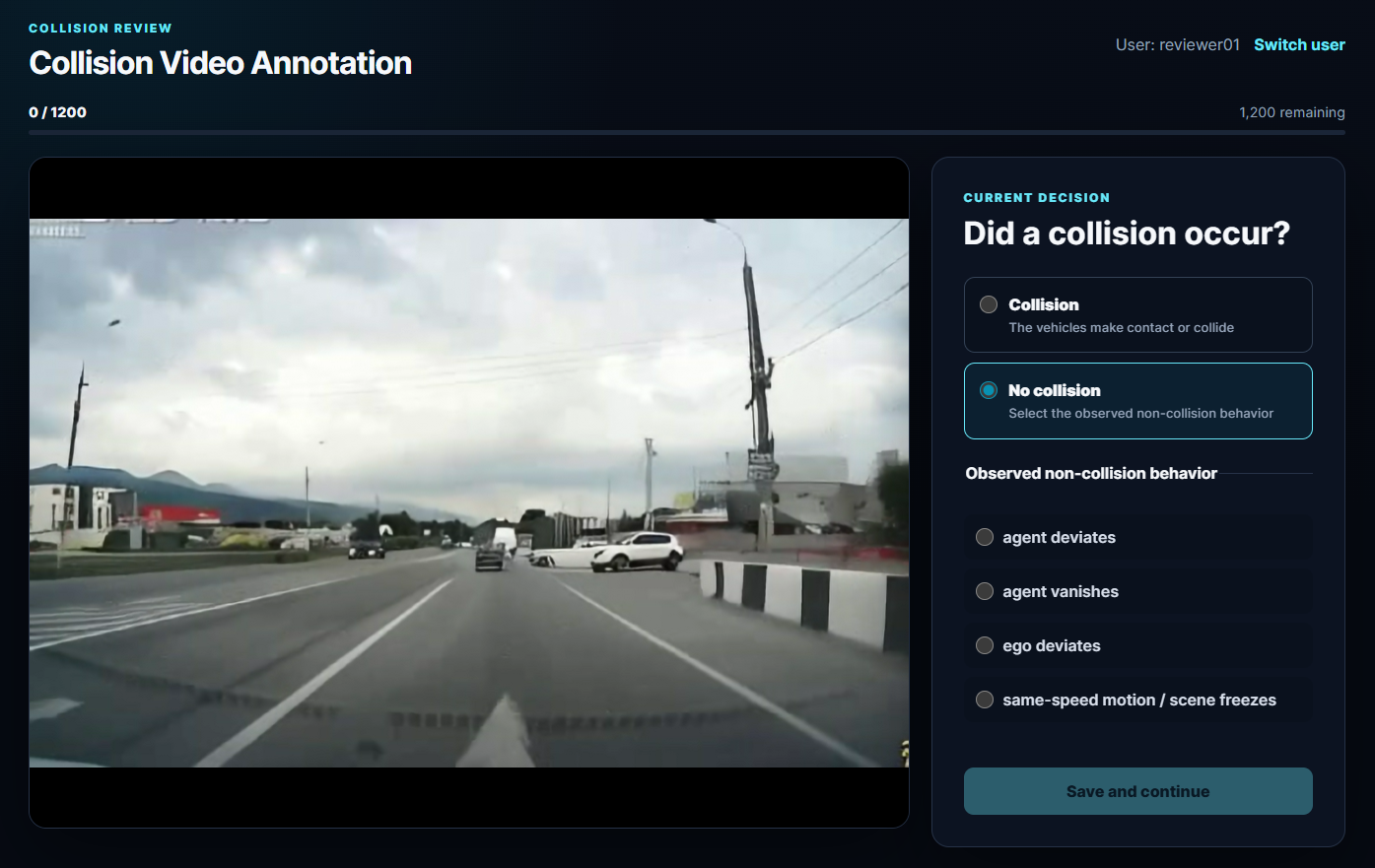}
  \caption{\textbf{Interface for manual assessment of generated outcomes.}
  The interface presents generated videos for inspection with model names
  hidden and the presentation order randomized. Collision outcomes are
  assessed over the complete 25-frame clip.}
  \label{sfig:annotation-ui}
\end{figure}

Generated videos are inspected using a dedicated annotation interface,
with model names hidden and presentation order randomized
(Supplementary Fig.~\ref{sfig:annotation-ui}). Each complete 25-frame clip
is assigned a collision or no-collision label based on the visible
interaction. At the 0.5-s prediction start, no-collision clips are
further assigned one of four qualitative avoidance modes according to the
apparent behavior: \emph{agent deviates}, when another road user changes
speed or direction to avoid the conflict; \emph{agent vanishes}, when that
road user disappears or fades from the generated scene; \emph{ego deviates},
when the generated ego motion departs from the conditioned trajectory; and
\emph{same-speed motion / scene freezes}, when the relative motion ceases
to close the gap or the scene stops evolving or degrades.

Let $C_{m,i,h}\in\{0,1\}$ denote the collision label for model $m$, source
video $i$ and prediction start $h$. Collision recall is
\begin{equation}
  \operatorname{Recall}_{m}(h)
  =\frac{1}{100}\sum_{i=1}^{100}C_{m,i,h}.
  \label{seq:collision-recall}
\end{equation}
The outcome-composition analysis at $h=0.5$\,s expresses both collision and
avoidance-mode counts as fractions of all 100 input scenes. The four
avoidance categories are assigned within the no-collision subset.

\section{Real-world experimental setup}\label{smeth:realworld}

\subsection{Research vehicle and image acquisition}
\begin{figure}[htbp]
  \centering
  \includegraphics[width=0.99\linewidth]{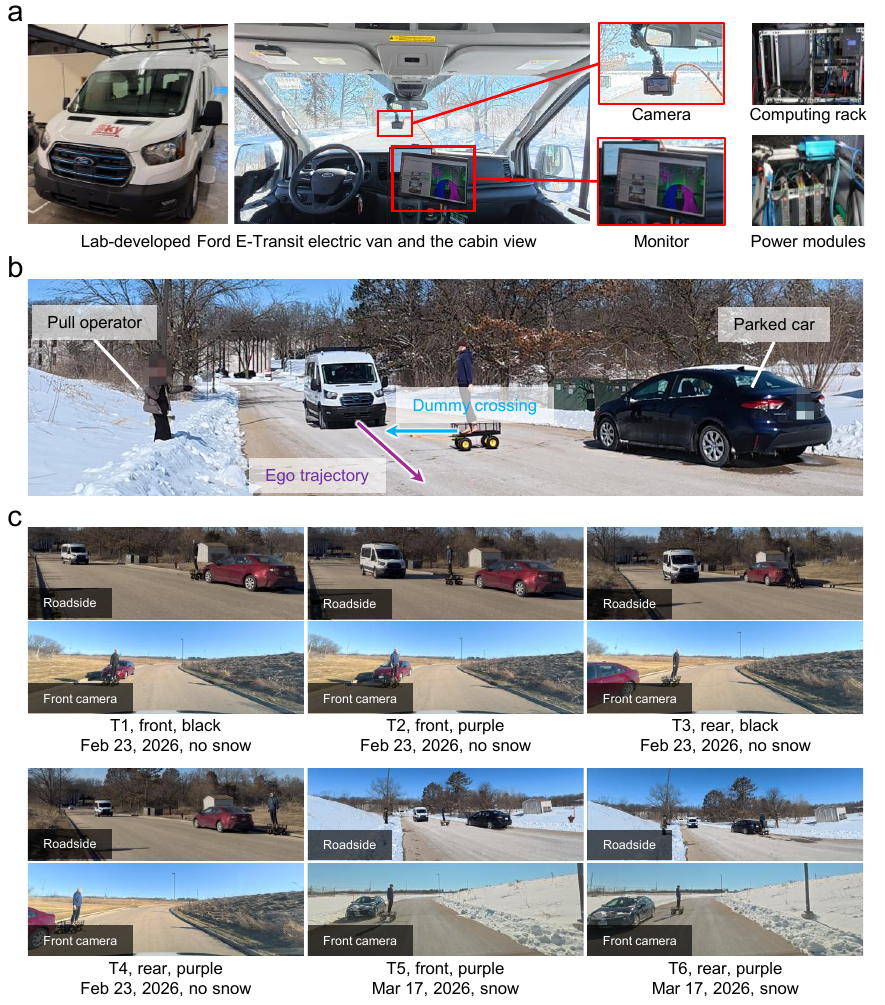}
  \caption{\textbf{Research van and real-world crossing configurations.}
  \textbf{a}, Ford E-Transit research van operated by the Wisconsin Transportation Center and sensors.
  \textbf{b}, Pedestrian dummy and wheeled cart used for manually pulled
  crossings. \textbf{c}, Recorded views of the six configurations in
  Supplementary Table~\ref{stab:realworld-trials}, showing variations in
  snow presence, clothing and crossing origin relative to a parked
  vehicle.}
  \label{sfig:realworld-platform}
\end{figure}

Field recordings were collected using a Ford E-Transit electric research
van equipped with a front-facing camera. Images were
cropped to $1600\times900$ pixels for storage in a nuScenes-style data
layout. The model-input pipeline subsequently resizes the complete stored
image to $512\times320$ pixels and normalizes RGB intensities to $[-1,1]$.
Supplementary Fig.~\ref{sfig:realworld-platform} illustrates the vehicle,
camera and crossing configurations.

\subsection{Recorded vehicle states and temporal alignment}
The recordings contain timestamped ego positions, vehicle speed,
throttle, a brake signal and steering-wheel angle. Image timestamps
are read from the stored filenames and matched to image references in the
driving log. For each recording, the image-to-log offset is estimated as
the median difference between an image's first appearance in the driving
log and its elapsed image timestamp. This offset maps images onto the
vehicle-log time axis. Position and speed are then linearly interpolated
at the corresponding times. Brake measurements are reported in their
original recorded units.

\subsection{Trial configurations and decision-frame selection}
Six pedestrian-crossing trials were recorded on a controlled route in
Fitchburg, Wisconsin, USA, on February 23 and March 17, 2026
(Supplementary Table~\ref{stab:realworld-trials}). A pedestrian dummy on a
wheeled cart was manually pulled across the vehicle's path. The
configurations varied snow presence, the dummy's clothing and whether
the crossing originated in front of or behind a parked vehicle. 
A human safety operator drove the research vehicle during data collection.

Decision frames were selected manually from the original recordings. Each trial contributes five \emph{Normal} frames and five \emph{Response} frames in temporal order within each stage. Normal frames show the vehicle approaching while the dummy has not begun to cross. Response frames show the vehicle responding to the crossing. Selection followed the visible crossing stage and recorded vehicle motion, with trial-specific decision times. The six trials therefore provide 60 decision frames, with 30 in each stage. Before model inference, the on-site test personnel visible in the scene were digitally masked out in the selected decision frames where necessary, so that their presence would not affect the imagined futures or the subsequent action selection. The same processed frames were used for all models.

\begin{table}[htbp]
  \centering
  \caption{\textbf{Configurations of the six real-world trials.}
  Crossing origin is specified relative to the parked vehicle. Snow
  presence describes the observed site conditions. Each trial contributes
  five Normal and five Response decision frames.}
  \label{stab:realworld-trials}
  \small
  \begin{tabularx}{\linewidth}{@{}clcXl@{}}
    \toprule
    Trial & Date & Snow present & Crossing origin & Dummy clothing \\
    \midrule
    T1 & February 23, 2026 & No  & In front of parked vehicle & Black \\
    T2 & February 23, 2026 & No  & In front of parked vehicle & Purple \\
    T3 & February 23, 2026 & No  & Behind parked vehicle     & Black \\
    T4 & February 23, 2026 & No  & Behind parked vehicle     & Purple \\
    T5 & March 17, 2026    & Yes & In front of parked vehicle & Purple \\
    T6 & March 17, 2026    & Yes & Behind parked vehicle     & Purple \\
    \bottomrule
  \end{tabularx}
\end{table}

\subsection{Vehicle dynamics of the recorded real-world trials}
Recorded displacement, speed, throttle and brake signals provide the
temporal context for the selected decision frames, as illustrated in
Supplementary Fig.~\ref{sfig:realworld-dynamics}. In each panel, time is
referenced to the first Normal frame (A1). Every recording contains an
approach, a deceleration and a stopped portion. The five Normal frames
(A1--A5) are taken from the approach, during which the throttle is applied
and the speed reaches approximately 2--4~m/s. The five Response frames
(B1--B5) lie on the falling portion of the speed trace after the throttle is released.

\begin{figure}[htbp]
  \centering
  \includegraphics[width=0.99\linewidth]{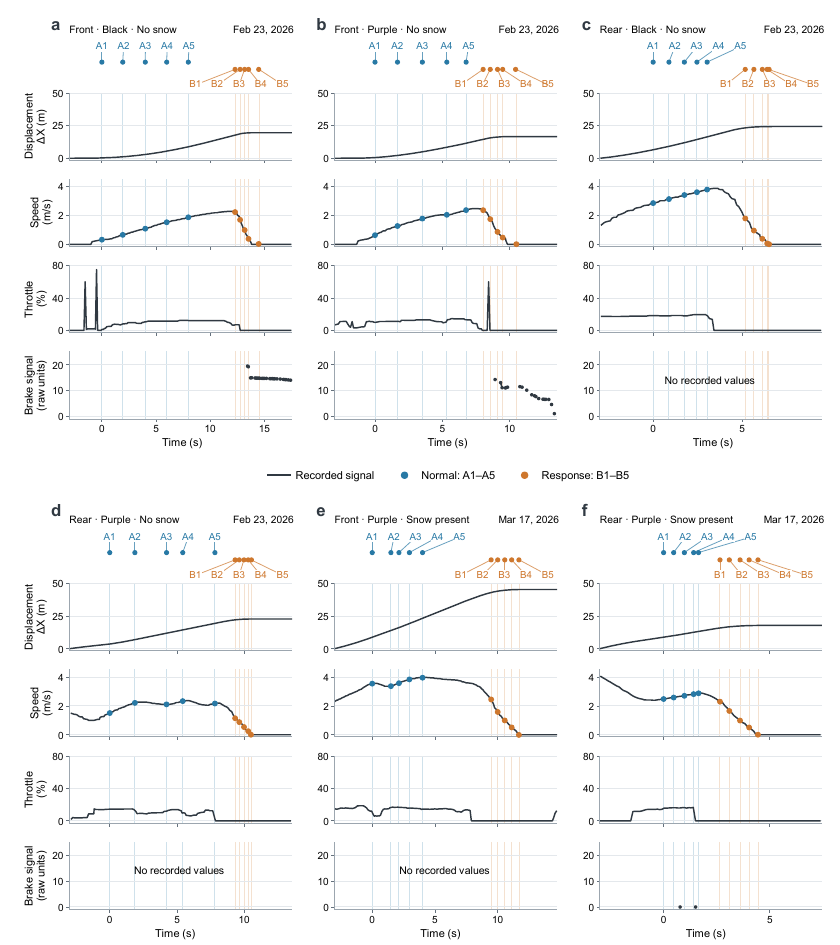}
  \caption{\textbf{Recorded vehicle motion and decision-frame selection.}
  Aligned displacement, speed, throttle and brake signals for the six
  trials. Markers identify the five Normal frames (A1--A5) and five
  Response frames (B1--B5) selected from each recording. 
  Brake signals are shown in their recorded raw units.}
  \label{sfig:realworld-dynamics}
\end{figure}

\section{Reward computation and world-model-guided action selection}
\label{smeth:reward}

\begin{figure}[htbp]
  \centering
  \includegraphics[width=0.99\linewidth]{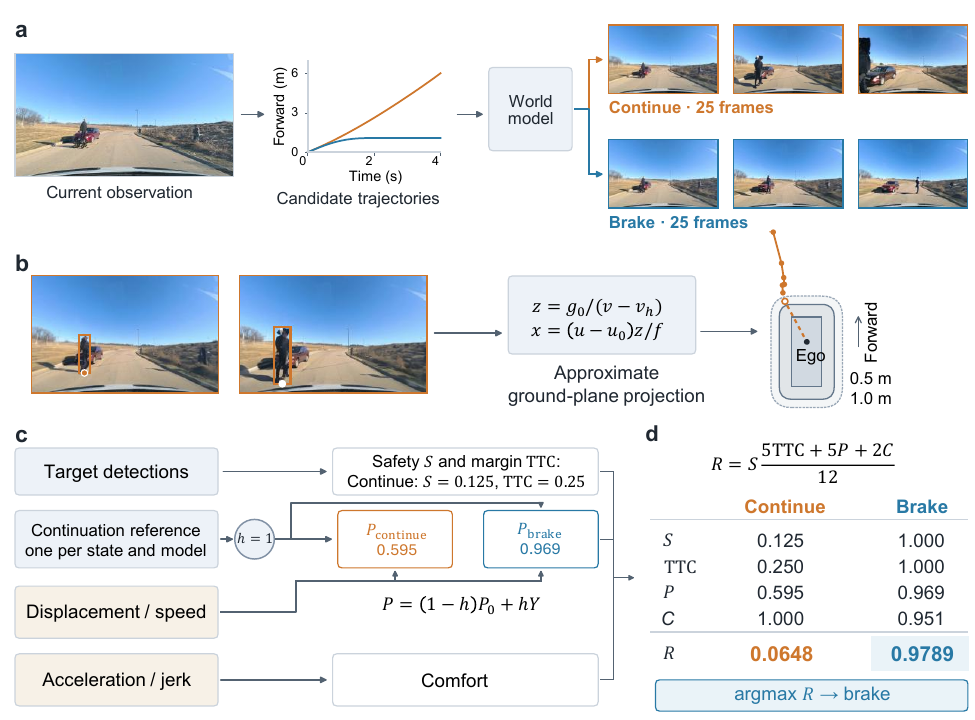}
  \caption{\textbf{Computation of reward from imagined futures.}
  \textbf{a}, Candidate trajectories generate alternative futures from a
  shared current observation. \textbf{b}, Detection and temporal
  association yield target boxes, which are projected to approximate
  ground-plane positions and vehicle clearances. \textbf{c}, Observed
  proximity and warning times determine $S$ and $\mathrm{TTC}$; reference
  context and candidate motion determine $P$ and $C$.
  \textbf{d}, The weighted reward ranks the candidate actions.}
  \label{sfig:reward-pipeline}
\end{figure}

For each candidate ego trajectory, the world model generates a 25-frame
future from the current observation. We evaluate this future using the
reward in Methods Section~4.2.2,
\begin{equation}
  R=S\,\frac{w_T\,\mathrm{TTC}+w_P P+w_C C}{w_T+w_P+w_C},
  \qquad (w_T,w_P,w_C)=(5,5,2).
  \label{seq:reward-aggregation}
\end{equation}
The four components are a safety gate $S$, a temporal safety-margin score
$\mathrm{TTC}$, a context-dependent progress score $P$, and a longitudinal
comfort score $C$, each in $[0,1]$. Supplementary
Fig.~\ref{sfig:reward-pipeline} summarizes the computation.
Video-derived detections and fixed camera geometry supply the safety cues; the known candidate trajectory supplies displacement, speed, acceleration and jerk. Traffic context is inferred from an imagined continuation of the current motion. No ground-truth future agent states or NAVSIM scoring calls enter the reward computation. The weights follow the temporal-margin, progress and comfort weights of the NAVSIM predictive driver model score (PDMS). Reward parameters were fixed before evaluation and shared across the model comparisons.

\subsection{Detection, temporal association and clearance estimation}
Generated frames are evaluated at $512\times320$ pixels. The timestamps span $t\in[0,H]$, with $H=4$\,s. 
We use YOLO11x to detect surrounding road users in the generated frames. Detected agents are associated across frames using bounding-box overlap, center displacement and detection confidence. The subsequent reward computation follows the same procedure in both settings.

\begin{table}[htbp]
  \centering
  \small
  \caption{\textbf{Reward parameters and detection settings.}
  Values for the reward implementation. IoU, intersection over union; NMS, non-maximum suppression.}
  \label{stab:reward-params}
  \begin{tabularx}{\linewidth}{@{}>{\raggedright\arraybackslash}p{0.25\linewidth}
    >{\raggedright\arraybackslash}X
    >{\raggedright\arraybackslash}p{0.27\linewidth}@{}}
    \toprule
    Parameter & Meaning & Value \\
    \midrule
    $w_T,w_P,w_C$ & Reward weights & $5,5,2$ \\
    $H$, $T_{\mathrm{safe}}$ & Rollout horizon; warning-time normalization & $4$ s; $4$ s \\
    $\alpha$ & Proximity-event gate coefficient & $0.5$ \\
    $d_c$, $d_w$ & Proximity and warning clearances & $0.5$ m; $1.0$ m \\
    $a_0$, $j_0$ & Acceleration and jerk scales & $3$ m\,/\,s$^{2}$; $5$ m\,/\,s$^{3}$ \\
    $\lambda_j$ & Relative jerk-cost weight & $1$ \\
    Detector & Checkpoint & YOLO11x \\
    Detection settings & Confidence; NMS IoU & $0.10$; $0.60$ \\
    Duplicate detections & IoU threshold for removing competing duplicates & $0.35$ \\
    \bottomrule
  \end{tabularx}
\end{table}

For a detected box $(x_1,y_1,x_2,y_2)$, its bottom center approximates the
target's ground contact point.
Coordinates are first mapped back to the corresponding source-image resolution \(W_{\mathrm{src}} \times H_{\mathrm{src}}\): $u=\tfrac{x_1+x_2}{2}(W_{\mathrm{src}}/512)$ and $v=y_2(H_{\mathrm{src}}/320)$.
An estimated planar camera model gives camera-relative lateral and
forward coordinates $(x,z)$:
\begin{equation}
  z=\frac{g_0}{v-v_h},
  \qquad x=\frac{(u-u_0)z}{f},
  \qquad v>v_h+1.
  \label{seq:reward-ground-projection}
\end{equation}
Here $v_h$ is the estimated horizon, $u_0$ is the forward vanishing-point coordinate, $f$ is the effective focal length, and $g_0$ is the ground-plane scale in pixel meters. 
Camera geometry is obtained from the dataset calibration for NAVSIM and estimated from the recorded vehicle motion for the real-world experiments.

The ego footprint is the rectangle
$\mathcal B=[-W/2,W/2]\times[d_f-L,d_f]$. With target radius $r_p$, the clearance proxy is
\begin{equation}
  \begin{aligned}
  \delta(t)&=\sqrt{d_x(t)^2+d_z(t)^2}-r_p,\\
  d_x(t)&=\max\{|x(t)|-W/2,0\},\\
  d_z(t)&=\max\{z(t)-d_f,\ d_f-L-z(t),\ 0\}.
  \end{aligned}
  \label{seq:reward-clearance}
\end{equation}
The same clearance is compared with the proximity and warning thresholds,
without adding either margin to the footprint a second time.

\subsection{Safety gate and temporal safety margin}
A proximity event is detected when $\delta(t)\leq d_c=0.5$\,m, and a
warning event when $\delta(t)\leq d_w=1.0$\,m. Between consecutive valid
observations separated by at most 0.18\,s, the projected position is
linearly interpolated to estimate the first threshold crossing. Intervals
containing a missing observation are not bridged. Let $t_c$ and $t_w$
denote the first proximity and warning times, respectively, with
$t_c=\infty$ or $t_w=\infty$ when the corresponding event is not observed.
The safety gate is
\begin{equation}
  S=\begin{cases}
    \alpha\,t_c/H, & 0\leq t_c\leq H,\\
    1, & t_c=\infty,
  \end{cases}
  \qquad \alpha=0.5.
  \label{seq:safety-gate}
\end{equation}
Earlier proximity events therefore suppress the reward more strongly.
The temporal safety-margin component is
\begin{equation}
  \mathrm{TTC}=\begin{cases}
    \min\{t_w/T_{\mathrm{safe}},1\}, & 0\leq t_w\leq H,\\
    1, & t_w=\infty,
  \end{cases}
  \qquad T_{\mathrm{safe}}=4\,\mathrm{s}.
  \label{seq:ttc}
\end{equation}
Thus, $\mathrm{TTC}$ denotes the normalized time to the first warning in
the imagined rollout. Both terms use video-derived proximity cues.

Near the vehicle, the target's ground-contact point can become occluded before contact is
visible. A near-field rule is activated when the box bottom reaches the
lowest 10\% of the image and its height is at least 25\% of image height.
The box bottom then bounds the possible ground-contact depth. If any
position along the corresponding ray, from a 0.3-m near plane to that
depth, intersects the proximity region, both a proximity and a warning
event are recorded at that frame. Other truncated or unprojectable boxes
are treated as unresolved observations. This rule supplies a near-field
risk proxy rather than a measured physical-contact time.

\subsection{Traffic context, progress and comfort}
For each initial observation, the world model evaluates a
continuing-motion reference rollout. The binary context $h$ indicates
whether this reference exhibits a warning or proximity event under the
same thresholds. The resulting context is shared by the candidate
trajectories evaluated from that observation. This procedure is applied
with the same rules to NWM and Vista, each using its own imagined
reference rollout.

Let $s(t)$ be displacement along the initial forward axis and $v(t)$ its
time derivative. With reference speed $v_{\mathrm{ref}}$,
define normalized forward progress $P_0$ and a yielding score $Y$ by
\begin{equation}
  \begin{aligned}
    P_0&=\operatorname{clip}\!\left(
        \frac{s(H)-s(0)}{v_{\mathrm{ref}}H},0,1\right),\\
    Y&=1-\frac{1}{H}\int_0^H
        \operatorname{clip}\!\left(\frac{v(t)}{v_{\mathrm{ref}}},0,1\right)
        \,\mathrm{d}t,\\
    P&=(1-h)P_0+hY.
  \end{aligned}
  \label{seq:progress}
\end{equation}
Progress uses net forward displacement. When the reference indicates a
need to yield, slower motion receives a higher $Y$; otherwise, the score
favors forward progress. The context-conditioned quantity $P$ here
corresponds to the \texttt{behavior\_score} in the implementation.

Longitudinal comfort is computed from the candidate-motion profile:
\begin{equation}
  C=\exp\!\left[-\frac{1}{H}\int_{\mathcal I_C}
       \left(\frac{a(t)^2}{a_0^2}
             +\lambda_j\frac{j(t)^2}{j_0^2}\right)\mathrm{d}t\right],
  \label{seq:comfort}
\end{equation}
where $a(t)=\dot v(t)$, $j(t)=\dot a(t)$ and $\mathcal I_C$ is the
scored portion of the motion profile. For analytic braking profiles, the squared-acceleration and squared-jerk integrals are evaluated analytically. The yielding integral uses the analytic speed profile or piecewise-linear sampled speed.

\subsection{Aggregation and candidate selection}
\begin{figure}[htbp]
  \centering
  \includegraphics[width=0.99\linewidth]{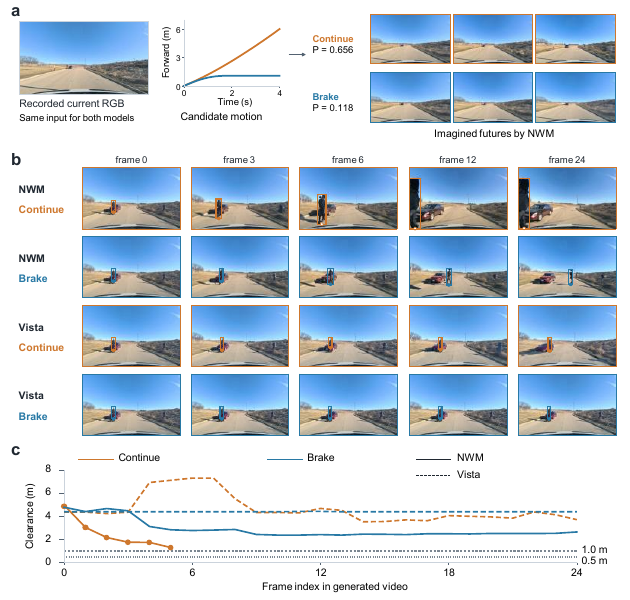}
  \caption{\textbf{Examples of reward-guided action selection.}
  \textbf{a}, A routine-driving input illustrates the selection for
  continuing motion when progress is appropriate. \textbf{b}, A
  pedestrian-crossing input shows candidate futures generated by
  NWM and Vista, with tracked target detections. \textbf{c}, Clearance
  estimates show how the safety components are obtained.}
  \label{sfig:reward-examples}
\end{figure}

The same reward expression and parameter values are used for both world
models. Each candidate is scored from one generated future. The selected
candidate is
\begin{equation}
  k^*=\underset{k}{\operatorname{arg\,max}}\;
       R(\hat x^{(k)},\tau^{(k)}).
  \label{seq:reward-action-selection}
\end{equation}
Supplementary Table~\ref{stab:reward-params} lists the implementation parameters. Supplementary Fig.~\ref{sfig:reward-examples} illustrates two initial
states, each evaluated with continuing-motion and braking candidates.
In the first example, progress scores are 0.656 and 0.118
for continuation and braking, respectively. In the second example, the
NWM continuation triggers the proximity rule at the 5th frame. Its reward is 0.0648, compared with
0.9789 for braking. For the same second input, Vista yields rewards of
0.7521 and 0.5878, respectively, and selects continuation. The figure
connects the detected interaction, component scores and final ranking.

\subsection{NAVSIM candidate selection and evaluation}\label{smeth:navsim}
ExploreVLA and AutoVLA provide two candidate trajectories for each scene.
Let $q_{i,E}$ and $q_{i,A}$ be their official NAVSIM PDMS values. The
selected evaluation set $\mathcal D$ contains two equally sized groups:
\begin{equation}
  \begin{aligned}
    \mathcal D_E&=\{i\in\mathcal D:q_{i,E}>0.9,\ q_{i,A}<0.2\},
       &|\mathcal D_E|&=100,\\
    \mathcal D_A&=\{i\in\mathcal D:q_{i,A}>0.9,\ q_{i,E}<0.2\},
       &|\mathcal D_A|&=100.
  \end{aligned}
  \label{seq:navsim-subset}
\end{equation}
These conditions define the 200 selected scenes and balance which policy
provides the successful candidate. Benchmark scores are used offline to
construct this subset and evaluate the selected trajectories. During
world-model-guided selection, candidates are ranked by the reward of their
imagined futures. Both world-model selectors receive the same candidate
pair and current observation for each scene.
For model $m$, let $\hat k_{i,m}\in\{E,A\}$ be its selected candidate.
The reported PDMS is the mean official score of the selected trajectories,
\begin{equation}
  \overline q_m=\frac{1}{200}\sum_{i=1}^{200}q_{i,\hat k_{i,m}},
  \qquad
  \overline q_{\mathrm{oracle}}
    =\frac{1}{200}\sum_{i=1}^{200}\max\{q_{i,E},q_{i,A}\}.
  \label{seq:navsim-selection-score}
\end{equation}
Individual-policy baselines always select the corresponding policy's candidate.

\end{document}